\documentclass[10pt,final]{IEEEtran}
\usepackage{mathtools}
\usepackage{graphicx}
\graphicspath{{figs/}}
\usepackage[utf8]{inputenc} 
\usepackage[T1]{fontenc}    
\usepackage[dvipsnames]{xcolor}         
\usepackage[noadjust]{cite}
\usepackage[bookmarks=true,breaklinks=true,colorlinks,linkcolor=RedViolet,citecolor=blue,urlcolor=BlueViolet]{hyperref}
\usepackage{url}            
\usepackage{booktabs}       
\usepackage{amsmath,amssymb,amsfonts,amsthm,stmaryrd}
\usepackage{nicefrac}       
\usepackage{microtype}      
\usepackage{authblk}
\usepackage[linesnumbered,ruled,vlined]{algorithm2e}
\usepackage{subcaption}
\usepackage{dblfloatfix}
\usepackage{enumitem} 
\usepackage{dsfont}
\usepackage[misc]{ifsym} 

\usepackage{siunitx}
\renewcommand{\eqref}[1]{Eq.\,(\ref{#1})}

\usepackage{booktabs,makecell, multirow, tabularx}

\newtheorem{definition}{Definition}

\usepackage{longtable}

\newcommand{\yA}[1]{\hat{y}_{\text{obs}}}
\newcommand{\yR}[1]{\hat{y}_{\text{rec}}}

\newcommand{\romup}[1]{\uppercase\expandafter{\romannumeral #1\relax}}
\newcommand{\romlo}[1]{\lowercase\expandafter{\romannumeral #1\relax}}

\newcommand{\rom}[1]{\uppercase\expandafter{\romannumeral #1\relax}}
\newcommand{\romsm}[1]{\lowercase\expandafter{\romannumeral #1\relax}}

\newcommand{\alg}{EdgeGFL}

\title{EdgeGFL: Rethinking Edge Information in Graph Feature Preference Learning}

\author{
  Shengda~Zhuo,~
  Jiwang~Fang,~
  Hongguang~Lin,~
  Yin Tang,~
  Min Chen, \IEEEmembership{Fellow, IEEE},~\\
  Changdong~Wang, \IEEEmembership{Senior Member, IEEE},~
  and~
  Shuqiang~Huang~
  \thanks{
    Shengda Zhuo, Jiwang Fang and Hongguang Lin are College of Cyber Security, Jinan University, Guangzhou 510632, China. E-mail: Zhuosd96@gmail.com, fjw2021@stu2021.jnu.edu.cn, hglin@stu2022.jnu.edu.cn
  }
  
  \thanks{
    Yin Tang is with the School of Management,
    Jinan University, Guangzhou, China. 
    E-mail: ytang@jnu.edu.cn
  }
  \thanks{
    Changdong Wang is with 
    with School of Computer, Sun Yat-sen University, Guangzhou 510275, China. E-mail: changdongwang@hotmail.com
  }
  \thanks{
    Min Chen is with the School of Computer Science and Engineering, 
    South China University of Technology, Guangzhou, China. 
    E-mail: minchen@ieee.org
  } 
  \thanks{
  Shuqiang Huang is with College of Cyber Security of Jinan University, Jinan University, 
  and also with Guangdong Key Laboratory of Data Security and Privacy Preserving, Guangzhou 510632, China. E-mail: hsq@jnu.edu.cn
  }
  \thanks{
    Corresponding Authors: Shuqiang Huang (hsq@jnu.edu.cn)
  }
}

\begin{document}
\pagenumbering{gobble}
\maketitle
\thispagestyle{plain}

\begin{abstract}
  Graph Neural Networks (GNNs) have significant advantages in handling non-Euclidean data and have been widely applied across various areas, thus receiving increasing attention in recent years.
  The framework of GNN models mainly includes the information propagation phase and the aggregation phase, treating nodes and edges as information entities and propagation channels, respectively. 
  However, most existing GNN models face the challenge of disconnection between node and edge feature information, as these models typically treat the learning of edge and node features as independent tasks. 
  To address this limitation, we aim to develop an edge-empowered graph feature preference learning framework that can capture edge embeddings to assist node embeddings. 
  By leveraging the learned multidimensional edge feature matrix, we construct multi-channel filters to more effectively capture accurate node features, thereby obtaining the non-local structural characteristics and fine-grained high-order node features. 
  Specifically, the inclusion of multidimensional edge information enhances the functionality and flexibility of the GNN model, enabling it to handle complex and diverse graph data more effectively. 
  Additionally, integrating relational representation learning into the message passing framework allows graph nodes to receive more useful information, thereby facilitating node representation learning. 
  Finally, experiments on four real-world heterogeneous graphs demonstrate the effectiveness of the proposed model.

\end{abstract}

\section{Introduction}
\label{sec:intro}

\IEEEPARstart{G}{raphs} generally contain rich node features and edge features. 
However, in recent years, most advanced GNN models have mainly focused on enhancing the learning of node features while neglecting the simultaneous learning of edge features. 
Although the aggregation functions designed based on the Message Passing Neural Network (MPNN) framework can aggregate both node features and edge features and have shown good results in specific application scenarios, using predefined aggregation functions is akin to manual feature engineering and cannot be generalized to all cases. 
Therefore, we aim to develop a method that can iteratively learn multidimensional edge features and synchronize the updates of these multidimensional edge features during the node information aggregation process. 
\begin{figure}[!t]
    \centering
    \includegraphics[width=0.41\textwidth]{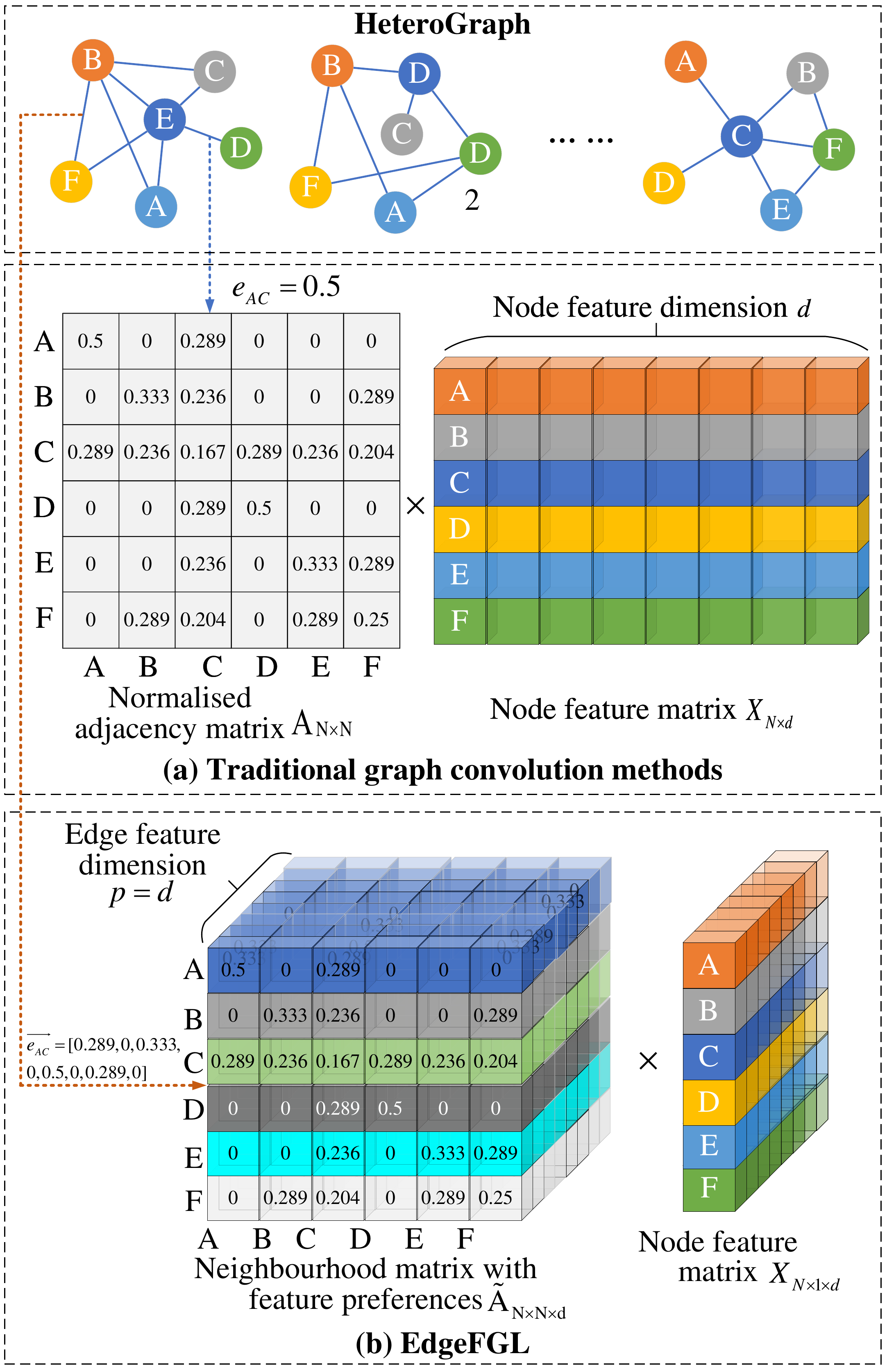}
    \caption{
    Illustration of the difference between information preferences and attention mechanisms in a message passing framework.
    \textbf{(a)} The edge feature representation in ordinary GCN;
    \textbf{(b)} The edge feature representation in our proposed \alg.
    }
    \label{fig:intro1}
\end{figure}
This approach fully leverages the edge features to assist in node representation learning. 
As shown in Fig.~\ref{fig:intro1}(a), edge features in traditional Graph Convolutional Networks (GCNs) are represented by an adjacency matrix, which can only be expressed with binary indicator variables or one-dimensional real values, failing to capture rich edge information. 
In contrast, Fig.~\ref{fig:intro1}(b) illustrates our proposed method for representing multidimensional edge features. 
Edge features are no longer represented by a single real value in the adjacency matrix but by learnable feature vectors, which can express rich edge information.


Many methods have been developed in learning node representations that can better reflect the local and global structure in graph \cite{caos2015learninggraphrepresentations, du2021tabularnet, yu2020structured}. 
However, the above models mainly adopt traditional machine learning methods such as random walk and matrix factorization. 
By considering the remarkable success of deep learning models in learning representations, many attempts have been made to utilize neural network models for network embedding \cite{wang2016structural, cao2016deep, he2019content}. 
These approaches mainly aim to adapt graph-structure data to the traditional neural networks designed for specific tasks, which may not ‘understand’ the inherent characteristic of graph \cite{velivckovic2017graph}. 
For instance, the effectiveness of Convolutional Neural Network (CNN) is based on its grid-like structure of image, and the expressive power of Recurrent Neural Network (RNN) \cite{lipton2015critical} is related to the sequential information within the corpus. 
However, none of them consider the topological relation of graph.


To solve this problem, Heterogeneous Graph Neural Network (HGNN) \cite{shi2022heterogeneous, wang2019heterogeneous} models have been developed. 
By considering each node as the information entity, and each edge as the propagation channel, many HGNN models apply a message-passing framework \cite{xu2022explicit, ding2021diffmg} to depict the training process, which is mainly composed of the \textbf{information propagation} phase and the \textbf{information aggregation} phase. 
In the information propagation phase, nodes send their information to their neighbors through the edge channel, where information may be modified according to the target node, and the information passed through edges is called message. 
In the information aggregation phase, nodes update their information by aggregating the received messages through a specifically designed function.
For instance, in Graph Convolutional Network (GCN) \cite{kipf2016semi, yu2022multiplex}, nodes first propagate their information to neighbors, and then aggregate the received information to update their representations by averaging. 
%
Essentially, the weights in Graph Attention Networks (GAT) \cite{velivckovic2017graph} are functions of node features, and the attention weight coefficients between two connected nodes are computed based on the feature vectors of these two nodes. This makes GAT more adaptable to the fusion of node features and structural features, leading to better performance. However, edges also contain rich information. Although existing GNN models have made significant progress in some aspects, they still have deficiencies in handling edge features.

Two challenges may be caused by the above phenomenon.
\emph{\textbf{First}}, as the link between a pair of nodes, the edge contains the interaction information between them, e.g., the similar feature interests between the connected nodes, and the translation relation between the start node and end node, which cannot be sufficiently expressed by a scalar value.
\emph{\textbf{Second}}, as the propagation channel, the edge should refine the sent information in a finer grain, which can significantly enhance the effectiveness of information exchange. From the perspective of the target node, only a part of the received information can be meaningful, while the remaining may have negative impacts, which can be considered as a kind of noise.
Therefore, we aim to develop a method that can fully leverage multidimensional edge features and synchronize the updates of these features during the node information aggregation process to further enhance the performance of GNN models. 
This approach will enable nodes to acquire more precise and comprehensive information while also learning edge embeddings to improve the representation of edge features. 
With this improvement, nodes will not only obtain richer feature information but also achieve better results in processing more complex and diverse graph data.

We propose using \emph{Edge-empowered Graph Feature Preference Learning} (\alg) to jointly embed edges and nodes to address the above challenges. 
In this paper, we abandon the traditional GCN method of using binary indicator variables or one-dimensional real values to represent edge features and instead introduce multidimensional edge embeddings to fully utilize edge information. 
We also propose a new message passing framework that integrates multidimensional edge features and node features, thereby fully leveraging both node and edge information.
In each training epoch of EdgeFGL, the process can be simply divided into the following steps. During the information propagation phase, each node sends its current representation to its neighbors through the connected edge channels. 
Specifically, each channel is represented as a relation vector, and when the node representation propagates through the channel, it is refined by a Hadamard product~\ref{horn1990hadamard}. 
Finally, each node receives the refined representation from the connected edges, and the node representation is updated through information aggregation. 
Integrating edge learning and node learning into the same convolutional layer greatly improves the model’s efficiency and reduces its complexity.

Experimental results show that our proposed method achieves state-of-the-art performance in both node classification and clustering tasks. 
The \textbf{contributions} of this paper are as follows:
\begin{itemize}[leftmargin=*]
    \item We propose a new message passing framework that can simultaneously aggregate multi- edge and node features. By introducing the concepts of residual connections and dense connections, we construct a deep graph convolutional neural network model to capture long-range dependencies and non-local structural features between nodes.
    \item We eliminate the limitation of traditional GNNs that only use binary variables or one-dimensional real values to represent edge features and propose a multidimensional edge feature representation method. Our approach uses edge embeddings to encode rich edge information, which can be iteratively updated across graph convolutional layers.
	\item We design a multidimensional edge feature matrix and construct multi-channel filters to filter node information, while introducing an identity mapping mechanism to prevent over-smoothing. Additionally, we conducted extensive experiments on four real-world datasets, verifying the superiority of our proposed model in node classification and clustering tasks compared to state-of-the-art baseline methods.
\end{itemize}

The subsequent sections of this paper are structured as follows:
In Sec. \ref{sec:Literature}, we briefly review related work. 
In Sec. \ref{sec:preliminary}, we introduce the preliminary knowledge and describe the motivation of this work.
In Sec. \ref{sec:method}, we describe in detail the proposed \alg~ model. 
In Sec. \ref{sec:experiments}, the experimental results are reported. 
Finally, the conclusion is drawn in Sec. \ref{sec:Conclude}.

\section{Related Work}
\label{sec:Literature}

This work relates to node representation learning, graph structure learning, and graph attention 
in training representation, so we review prior studies in each category and discuss the differences and relations.

\par\smallskip\noindent
\textbf{Node Representation Learning.}
Traditional node representation learning methods mainly focus on mapping the original node representation vector (e.g. the adjacency vector) into the continuous low-dimension embedding space \cite{goldberg2014word2vec}. 
Those works usually extend the classical machine learning models for graph-structured data. 
The early works like Locally Linear Embedding (LLE) \cite{roweis2000nonlinear}, Isomap \cite{tenenbaum2000global} and Laplacian Eigenmaps \cite{belkin2001laplacian} try to adopt the eigen-based methods, so as to keep the spectral properties of node when mapping into the lower space, which is time consuming and uneasy to be parallelized. 
Recently, inspired by word2vec \cite{mikolov2013distributed}, DeepWalk \cite{perozzi2014deepwalk} adopts random walk to generate node sequences to imitate sentences in corpus, and then feeds them into the Skip-gram framework [26] to train node representations, which can to some extent learn both local and global structure information.
Node2vec \cite{grover2016node2vec} improves the DeepWalk model by introducing a bias mechanism into the random walk, which can get more efficient node sequences to enhance the model performance. 
Besides, Modularized Nonnegative Matrix Factorization (M-NMF) \cite{wang2017community} focuses on extracting the mesoscopic community structure among nodes. 
Furthermore, Attributed Network Embedding with Micromeso structure (ANEM) \cite{li2021attributed} considers three kinds of information in node representation learning: attribute information, the local proximity structure, and mesoscopic community structure. 
As we can see, the above methods usually pay major attention to manually integrating special rules with shallow models.

\par\smallskip\noindent
\textbf{Graph Structure Learning.}
Due to the success of deep learning models in training representations, many attempts have been made to integrate them with graph-structured data. 
For example, Structural Deep Network Embedding (SDNE)~\cite{wang2016structural} uses deep auto-encoders to extract low-dimensional node representations and adds a pairwise loss to ensure connected nodes are close. 
Deep Neural Networks for Graph Representations (DNGR)~\cite{cao2016deep} calculates the Positive Point Mutual Information (PPMI)~\cite{bullinaria2007extracting} matrix for graphs and feeds it into a stacked denoising autoencoder to extract node representations. 
ContextAware Network Embedding (CANE)~\cite{tu2017cane} combines a mutual attention mechanism with CNNs to learn context-aware node representations. 
Self-Translation Network Embedding (STNE)~\cite{liu2018content} uses bi-LSTM on node sequences obtained by random walks to preserve node order information. 
Deep Attributed Network Embedding (DANE)~\cite{gao2018deep} assumes node representations from structure and feature information should be similar. 
Self-Paced Network Embedding (SPNE)~\cite{gao2018self} adopts a self-paced learning mechanism to learn node representations, considering information from neighbors of increasing hops. 
Deep Network Embedding with Structural Balance Preservation (DNE-SBP)~\cite{shen2018deep} uses stacked auto-encoders to preserve structural balance in signed networks. 
Nodepair Information Preserving Network Embedding (NINE)~\cite{wang2020node} uses adversarial networks to preserve local information between node pairs. 
However, traditional approaches often apply existing models to train node representations without specifically targeting graph-structured data.

\par\smallskip\noindent
\textbf{Graph Attention Learning.} 
The message-passing framework, designed to fit the topological characteristics of graphs, has been widely adopted in various graph neural networks. 
In Graph Convolutional Networks (GCN) \cite{kipf2016semi}, nodes propagate their information to neighbors and then update their representations by averaging the received information. 
Graph Attention Networks (GAT) \cite{velivckovic2017graph} improve on this by using an attention mechanism to assign different weights to different neighbors. 
Building on GCN and GAT, several methods have been developed to enhance the message-passing framework: Hierarchical Graph Convolutional Networks (HGCN) \cite{hu2019hierarchical} use a coarsening-refining model to hierarchically aggregate node information; 
Relation-aware co-attentive Graph Convolutional Networks (RecoGCN) \cite{xu2019relation} integrate a co-attention mechanism to adapt GCN for recommender systems by bridging user, item, and agent relations; 
Dynamic Hypergraph Neural Networks (DHNN) \cite{jiang2019dynamic} introduce a hypergraph mechanism to capture group relations, where each hyperedge covers a group of nodes; 
and Graphsage \cite{hamilton2017inductive} extends GCN to inductive learning and explores various aggregation operations like max pooling and LSTM. 
However, these models often treat each edge as a scalar, limiting edge representation capability. 
To better utilize edge information, some models incorporate edge representation learning: 
CensNet \cite{pan2019learning} constructs two graphs to represent node and edge adjacency, using shared trainable weights to link node and edge representations. 
Inspired by translation mechanisms \cite{jiang2019censnet}, edge representations are learned by assuming a transforming relation between edges and their corresponding nodes. 
Vectorized Relational Graph Convolutional Network (VR-GCN) \cite{ye2019vectorized} combines translation mechanisms with GCN to train node and relation representations in multi-relation network alignment tasks.

However, those models treat the representation learning of edge as the representation learning of node, without considering node interaction, and thus fail to learn the relation between the connected nodes. 
In this case, these models cannot properly integrate the edge representation learning into the message-passing framework.

\section{Background and moviation}
\label{sec:preliminary}

In this section, we begin by introducing the preliminaries that include the notations, preliminary (\emph{i.e.}, message-passing framework), and motivation. 

\subsection{Notations} \label{sec:notation}
Let $\mathcal{G} =\{\mathcal{V}, \mathcal{E}, \mathbf{F}, \phi, \psi \}$ denote an undirected attributed network, where $\mathcal{V} = \{ \upsilon_1, \upsilon_2, \cdots, \upsilon_{n} \}$ is a set of $n$ nodes, $\mathcal{E} = \{e_{ij}\ |\ i, j \in \text{nodes}, i \neq j\}$ is a set of $m$ edges, and a node attribute matrix $\mathbf{F} = [\mathbf{f}_{1};\ \cdots;\ \mathbf{f}_{n}] \in \mathbb{R}^{n \times d} $ with each row vector $\mathbf{f}_{i} \in \mathbb{R}^{n \times d} $ being the attribute vector of node $\upsilon_i$.
The $\phi(v)$ denotes the type of each node $\phi$, and $\psi(e)$ denotes the type of each edge $\psi$. 
Additional, each dimension being considered as a feature.
%
The collections of potential node types and edge types are symbolized as $\mathbf{T}_v=\{\phi(v): \forall v \in \mathcal{V}\}$ and $\mathbf{T}_e=\{\phi(e): \forall e \in \mathcal{E}\}$, correspondingly. 
When both $|\mathbf{T}_v|$ and $|\mathbf{T}_e|$ equal 1, the graph reverts to a conventional homogeneous graph.
Besides, each node $\upsilon_i \in \mathcal{V}$ has neighbor nodes, the index set of which is denoted as $\mathcal{N}_{i}$. 

We use $\mathbf{H}^{l} = [\mathbf{h}^{l}_{1}; \cdots; \mathbf{h}{^{l}_{n}}] \in \mathbb{R}^{n \times {d^{l}} }$ to denote the node representation matrix in the $l$-th layer of \alg~where the $i$-th row vector $\mathbf{h}^{l}_{i} \in \mathbb{R}^{n \times {d^{l}} }$ denotes the node representation vector of the $i$-th node in the $l$-th layer with dimensionality being $d^{l}$. 
Note that the initial node representation matrix is assigned as the node attribute matrix in the initial status, \emph{i.e.}, $\mathbf{H}^{l} = \mathbf{F}$ with $d^{0} = d$. 
The relation representation of each edge $e_{ij}$ in the $l$-th layer is denoted as $\mathbf{r}{^{l}_{ij}} \in \mathbb{R}^{n \times {d^{l}}}$. 
Besides, $\mathbf{Y} = [ y_{ik} ] \in \mathbb{R}^{n \times {c}}$ represents the label indicator matrix, where each element $y_{ik}$ indicates whether the $i$-th node has the $k$-th label.
Key notations used in the paper and their definitions are summarized in Table \ref{tab:notation}.
\begin{table}[!h]
	\centering
	\small
	\caption{Notations and Definitions.}
    	\begin{tabular}{l|p{5.5cm}}              
    		\toprule
    		\midrule
    		Notation    & Definitions  \\ 
    		\midrule
    		$\mathcal{G}=\{\mathcal{V},\mathcal{E},\mathbf{F}, \phi, \psi\}$    & The undirected attributed network   \\
                \midrule
                $\mathcal{V}=\{v_1, \cdots, v_n\}$    & The node set consisting of $n$ nodes  \\ 
                \midrule
                $\mathcal{E}=\{e_{12}, \cdots, e_{ij}\}$    & The edge set consisting of $m$ edges\\
                \midrule
                $\mathbf{F}=[ \mathbf{f}_{1}, \cdots, \mathbf{f}_{n} ]$  & The node attribute matrix \\
                \midrule
                $\mathbf{f}_{i} \in \mathbb{R}^{1 \times d}$ & The attribute vector of node $v_{i}$ \\
                \midrule
                $\mathbf{T}_{v}, \mathbf{T}_{e}$ & The sets of the possible node and edge types \\
                \midrule
                $\mathcal{N}_{i}$& The index set of neighbor nodes of $v_{i}$ \\
                \midrule
                $\sigma(\cdot)$ & The non-linear activation function. \\
                \midrule
                $\mathcal{S}(\cdot, \cdot)$ & The information propagation function.\\
                \midrule
                $\boldsymbol{o}_i$ & The output embedding of node $i$ \\
                \midrule
                $y_{ik}$ & The label indicator denoting whether the $i$-th node has the $k$-th label \\
                \midrule
                $\mathbf{H}^{l} = [h_{1}^{l}, \cdots, h_{n}^{l}]$ & The node representation matrix in the $l$-th layer\\
                \midrule
                $\mathbf{h}_{i}^{l} \in \mathbb{R}^{1 \times d}$  & The node representation vector of node $v_{i}$ in the $l$-th layer \\
                \midrule
                 $\mathbf{r}_{ij}^{l} \in \mathbb{R}^{1 \times d^{l}}$  & The relation representation of edge $e_{ij}$ in the $l$-th layer \\
                \midrule
                $\mathbf{Y}=[y_{ik}] \in \mathbb{R}^{n \times c}$ & The label indicator matrix\\
                \midrule
                $\mathbf{M}^{l}_{ij}\in \mathbb{R}^{1 \times d^{l}}$ & The representation of the sent message from node $v_i$ to node $v_j$ in the $l$-th layer\\
                \midrule
                $\mathbf{M}^{l}_{*i}\in \mathbb{R}^{1 \times d^{l}}$ & All of the information representations that node $v_i$ receives in the $l$-th layer\\
                \midrule
                $\mathbf{W}^{l} \in \mathbb{R}^{d^{1} \times d^{l+1}}$ & The trainable projection matrix in information propagation \\
    		\bottomrule
    	\end{tabular}
	\label{tab:notation}
\end{table}

\subsection{Message-passing Framework} \label{sec:message}
The message-passing framework is composed of information propagation and information aggregation phases.
The information propagation phase is formalized as follows:
\begin{eqnarray}
    \mathbf{M}_{j i}^l=\mathcal{S}\left(\mathbf{h}_j^l, e_{j i}\right),
    \label{eq:1}
\end{eqnarray}
where $\mathbf{M}_{ji}^l \in \mathbb{R}^{1 \times d^{l}}$ denotes the representation of the sent message from node $\upsilon_{j}$ to node $\upsilon_{i}$ in the $l$-th layer; $\mathcal{S}(\cdot, \cdot)$ denotes the information propagation function.

The information aggregation phase is formalized as follows:
\begin{equation}
    \mathbf{h}_i^{l+1}=\mathcal{R}\left(\mathbf{h}_i^l, \mathbf{M}_{* i}^l\right),
    \label{eq:2}
\end{equation}
where $\mathbf{M}_{* i}^l \in \mathbb{R}^{n \times d^{l}}$ denotes all of the information representations that node $\upsilon_{i}$ receives in the $l$-th layer;
$\mathcal{R}(\cdot, \cdot)$ denotes the information aggregation function. 
The two famous GNN-based models can be formalized via the above message-passing framework, as will be elaborated below.

1) \emph{Graph Convolutional Network (GCN)}: GCN can be reconsidered in the message-passing framework.
In the information propagation phase, the information propagation of GCN is specified as follows:
\begin{equation}
    \mathbf{M}_{j i}^l=a_{j i} \cdot \mathbf{h}_j^l,
    \label{eq:3}
\end{equation}
where $a_{ij}$ is the averaging factor computed by $\frac{1}{\sqrt{\left|\mathcal{N}_i\right| \cdot\left|\mathcal{N}_j\right|}}$, which is inversely proportional to the number of neighbors.

In the information aggregation phase, the aggregation function is constructed as follows:
\begin{equation}
    \mathbf{h}_i^{l+1}=\sigma\left(\sum_{j \in \mathcal{N}_i \cup i} \mathbf{M}_{j i}^l \cdot \mathbf{W}^l\right),
    \label{eq:4}
\end{equation}
where $\mathbf{W}^l \in \mathbb{R}^{n \times {d^{l}}}$ is the trainable projection matrix from the $l$-th layer, and $\sigma(\cdot)$ is the non-linear activation function.

2) \emph{Graph Attention Network (GAT)}: GAT improves the GCN model by introducing the attention mechanism to compute adaptive weights for different nodes in a neighborhood during the information propagation phase. 
That is, the information propagation phase is formalized as:
\begin{equation}
    \mathbf{M}_{j i}^l= \text{att} (\mathbf{h}_j^l, \mathbf{h}_i^l ) \cdot \mathbf{h}_j^l,
    \label{eq:5}
\end{equation}
where $\text{att}(\cdot, \cdot)$ is the attention function. 
The aggregation phase of GAT is the same as that of GCN, which can also be formalized as Eq. (\ref{eq:4}).

\subsection{Motivation.}
The goal of the message-passing framework is to refine the information of each node by exchanging their information with neighbors, which efficiently utilizes the topological relation inside graph. 
In this case, the effectiveness of information exchange is vital in the message-passing framework. 
As we can see from Eqs. (\ref{eq:3}) and (\ref{eq:5}), both GCN and GAT pay much attention to modifying the scale of information,
but lack the consideration of the feature preference of each node. 
\textbf{1)} Nodes usually prefer to receive the part of related features instead of the whole information. 
\textbf{2)} The unrelated features inside the received information may act as noise to the target node.

In this work, we attempt to develop a method to enhance the effectiveness of the message-passing framework, which is aware of the node feature preference during the information exchange process.

\begin{definition}[Feature Preference Aware Message-passing, FPAMP]
    When node $v_{i}$ sends information to neighbor $v_{j}$, instead of modifying the scale of whole information, the connected edge $e_{ij}$ can magnify the related features while shrinking the unrelated features.
\end{definition}

Toward this end, we integrate the edge representation into the message-passing framework and propose the \emph{Edge-Enhanced Graph Feature Preference Learning} (\alg) model in the next section, which can effectively enhance the performance of information exchange.




\section{The Proposed Approach}
\label{sec:method}

\par\smallskip\noindent
\subsection{Overview}
We now illustrate the proposed \emph{Edge-Enhanced Graph Feature Preference Learning} (\alg) model, which enhances the effectiveness of the information exchange procedure by introducing the edge representation as relation representation. 
Following the structure of the message-passing framework, each layer of \alg\ is mainly composed of two phases, namely Information Propagation and Information Aggregation.
    
\textbf{Information Propagation}: Different from the existing models (\emph{i.e.}, GCN and GAT), which only send information from one node to another in the information propagation phase, \alg\ first extracts the relation representation for edges from the corresponding node representations. 
Subsequently, each node sends the self-information to its neighbors, and features inside the sent information are magnified or shrunk through the corresponding edges, which finally forms a different ‘sent message’ according to the characteristics of the target nodes.

\textbf{Information Aggregation}: Upon receiving feature-preferred information from its neighbors, each node proceeds to refine its representation by integrating the accumulated messages.
Similarly, each edge receives the feature-preferred information from neighbors, and then, the node updates its representation by summing up the received message.

\begin{figure*}[!t]
    \centering
    \includegraphics[width=0.9\textwidth]{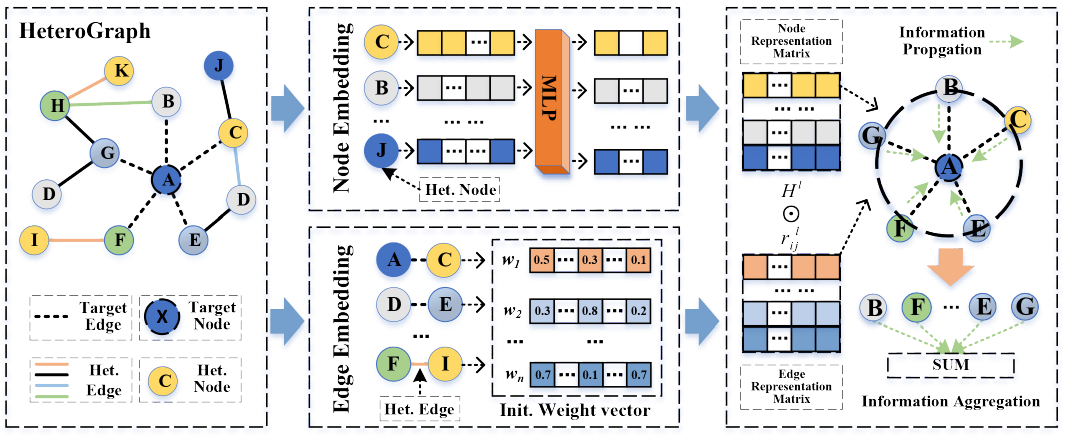}
    \caption{Illustration of the $l$-th layer of the \alg~model. The \alg~model achieves the final representation of the target node in a heterogeneous graph through node and edge embeddings, information propagation, and aggregation processes.}
    \label{fig:intro}
\end{figure*}

In general, each layer of the \alg~model is illustrated in Fig. \ref{fig:intro}. 
For each edge $e_{ij}$, there is a corresponding vectorized relation representation $\mathbf{r}_{ij}$ derived from the interaction between the connected nodes by using an initialized weight vector $\mathbf{W}(\mathbf{h}_{i}, \mathbf{h}_{j})$, and then each node sends information refined by the relation representation. 
Finally, the representation of each node is updated by integrating all of the refined information.

\subsection{Feature Projection} 

{\bf Node Nonlinear Transformation.} 
As the input features of different node types may differ in dimensionality, we apply a linear layer with bias for each node type and then utilize a non-linear mapping (\emph{e.g.}, using a rule or softmax function) through the activation function to map all node features to a shared feature space.
Typically, we use a learnable type transformation matrix $\mathbf{W}^{\mathbf{T}_{v}}$, where $\mathbf{T}_{v}$ is the set of node types, to map nodes of different types to the same dimensional space $\mathbf{H}^{l} = \mathbf{W}^{\mathbf{T}_{v}} * \mathbf{F}$.
To better map different types of nodes to the same semantic space, we apply a nonlinear activation function after a fully connected layer, as follows:
\begin{equation}
    \mathbf{H}^{l} = \text{LeakyRelu}(\mathbf{W}^{\mathbf{T}_{v}} * \mathbf{F} + b),
    \label{eq:6}
\end{equation}
where $\mathbf{F} = [ \mathbf{f}_1, \dots, \mathbf{f}_n ]$ is the attribute matrix of the node, and $\mathbf{H}^{l}=[\mathbf{h}^{l}_1, \dots, \mathbf{h}^{l}_n]$ is the representation matrix of the node.

{\bf Edge Type Initiation.} 
For heterogeneous graph $G$ with $\mathbf{T}_{e}=\{\phi(e): \forall e \in \mathcal{E}\}$ types of edges, an edge type dictionary $\mathbf{D}=\{E_1, E_2, \cdots, E_n\}$ is constructed by looking at the types of edges as different tokens and assigning an initialisation encoding vector $\mathbb{R}=\{\mathbf{r}_1,\mathbf{r}_2, \cdots ,\mathbf{r}_n\}$ of the corresponding type to the edges indexed by the different types based on the edge type dictionary. 
This vectorised representation holds the semantic information of the edge types:
\begin{align}
    \hat{\mathbf{r}}_{i j}^{l} &= f\left(\mathbf{T}_e=\{\phi(e): v_e \in \mathcal{E}\}\right),
    \label{eq:7}
\end{align}
where $f(.)$ is the initialization function for edge types, taking the values of each edge as input and producing corresponding edge vectors.

\subsection{Information Propagation} \label{sec:IP}
The edge type-based feature preference representation makes full use of the type information of the edges in the dataset by mapping the edge type information to the same data space as the node feature information, and by co-training with the node features in the same data space. 
The resulting feature preference representation can be used to scale each dimensional feature of the node information, rather than simply scaling all dimensional features with a fixed scalar.

We map the initialized coding vector $R$ to the data dimension space of the nodes via a learnable feature transformation matrix $w^{l}$, providing the basis for the co-training of node and edge features. 
The preference representation step for edge features can be expressed as:
\begin{align}
    \mathbf{r}_{i j}^l &= \delta(\hat{\mathbf{r}}_{i j} \cdot w^l),
    \label{eq:8}
\end{align}
where $\delta(\cdot)$ denotes the learnable initialization encoding function that assigns a corresponding initialisation encoding vector $\mathbf{r}$ to each edge type based on a dictionary of edge types, and $w$ is a learnable parameter shared by all types of edges that are used to map the encoding vectors of the edge types to the node feature space.

We apply edge feature preferences to a message-passing framework to replace aggregation weights in the form of scalars learned in convolutional and attentional neural networks. 
We propose to use learnable edge feature preferences to act as aggregation scales that both scale the message's size and learn the features' preferences. 
Specifically, the message-passing phase can be represented as:
\begin{equation}
    \mathbf{M}_{ij}^l = s\left(\mathbf{h}_j^l, \mathbf{r}_{ij}^l\right)=\mathbf{h}_j^l \cdot \mathbf{r}_{ij}^l,
    \label{eq:9}
\end{equation}
where $\mathbf{M}_{ij}^l$ denotes the message sent from node $v_j$ to node $v_i$ at layer $l$. $s(\cdot)$ denotes the message transfer function, and $\cdot$ denotes the dot product.

\subsection{Information Aggregation} \label{sec:IA}
In this phase, nodes update their representation by aggregating the received information via the sum aggregator, which can be formalized as:
\begin{equation}
    \mathbf{h}_i^{l+1} \leftarrow \underset{j \in \mathcal{N}_i}{\operatorname{SUM}_{ij}}(\mathbf{M}_{ij}^l).
    \label{eq:sum}
\end{equation}
By giving different weights to the target node and neighbors, we implement two kinds of sum aggregators as follows:
\begin{itemize}
    \item[1).]{{\bf Node Residual.} For the creation of node representations that span across layers, we introduce a pre-activation residual connection. The aggregation for the $l$-th layer can be formulated as:
    \begin{equation}
        \mathbf{h}_i^{l+1}=\sigma\left(\sum_{j \in \mathcal{N}_i} \alpha_{ij}^{l} \mathbf{W}^{l} \mathbf{h}_j^{l}+\mathbf{h}_i^{l}\right),
       \label{eq:node1}
    \end{equation}
    where, $\alpha_{ij}^l$ represents the attention weight corresponding to edge $\langle i, j\rangle $ for the $l$-th layer, and $\mathbf{W}^l \in \mathbb{R}^{n \times {d^{l}}}$ is the trainable projection matrix from the $l$-th layer. When there's a dimension alteration in the $l$-th layer, an extra trainable linear transformation $\mathbf{W}_{\text{res}}^{(l)} \in \mathbb{R}^{d^{l+1} \times d^{l}}$ becomes necessary. This can be expressed as:
    \begin{equation}
        \mathbf{h}_i^{l+1}=\sigma(\sum_{j \in \mathcal{N}_i} \alpha_{i j}^{l} \mathbf{W}^{l} \mathbf{h}_j^{l}+\mathbf{W}_{\text {res }}^{l} \mathbf{h}_i^{l}),
        \label{eq:node2}
    \end{equation}
    }

    \item[2)]{{\bf Edge Residual.} Recently, Realformer~\cite{he2020realformer} illustrated the advantageous nature of applying residual connections to attention scores. we incorporate residual connections into these scores:
    \begin{equation}
        \alpha_{i j}^{(l)}=(1-\beta) \hat{\alpha}_{i j}^{(l)}+\beta \alpha_{i j}^{(l-1)},
        \label{eq:edge}
    \end{equation}
    where hyperparameter $\beta \in [0, 1]$ is a scaling factor, $\hat{\alpha}$ employs both edge type embeddings and node embeddings to compute the attention score.
    }
\end{itemize}

\subsection{Model Training} \label{sec:MT}
Having aligned the final dimension of HGNNs with the class count, we employ standard loss functions. 
For \emph{single-label} classification, softmax and cross-entropy loss are used. 
For \emph{multi-label} datasets like IMDB in \alg, we apply sigmoid activation and binary cross-entropy loss. 
Notably, we observe the high utility of $L_{2}$ normalization on the output embedding.
\begin{equation}
{o}_i=\frac{\mathbf{h}_i^{L}}{\left\|\mathbf{h}_i^{L}\right\|},
\label{eq:normalization}
\end{equation}
where ${o}_i$ signifies the output embedding of node $i$. Normalization on the output embedding is a prevalent practice in retrieval-based tasks, as it ensures that the dot product corresponds to cosine similarity subsequent to normalization.

In the output layer, the predicted vector $\mathbf{z}_{i} \in \mathbb{R}^{1 \times c}$ of node $v_{i}$ is obtained from ${o}_i$ as follows:
\begin{equation}
    \mathbf{z}_i=\operatorname{Softmax}\left({o}_i \cdot \mathbf{W}_z\right) 
    \label{eq:z},
\end{equation}
where $\mathbf{W}_{z} \in \mathbb{R}^{d^{L \times c}}$ maps node representation to $c$-dimensional vector, which is normalized by Softmax(·) to make each dimension indicate the probability of the corresponding label.

After obtaining the probability matrix $\mathbf{Z} = [\mathbf{z}_{1}; \mathbf{z}_{2}; \dots , \mathbf{z}_{n}] \in \mathbb{R}^{n \times c}$, by computing the cross-entropy loss of the training data, the loss function of \alg\ with Similar Relation Learning is as follows: 
\begin{equation}
\mathcal{L}=-\sum_{i \in \mathcal{V}_{t r a i n}} \mathbf{Y}_{i *} \cdot \log \left(\mathbf{z}_i^{\top}\right) +\lambda\|\Theta\|_2,
\label{eq:loss}
\end{equation}
where $\mathcal{V}_{t r a i n}$ denotes the index set of nodes in the training set, $\mathbf{Y}_{i *} \in \mathbb{R}^{1 \times c}$ is the $i$-th row vector of the indicator matrix $\mathbf{Y}$ denoting the label vector of node $v_{i}$, and $\|\Theta\|_2$ is a regularization term of model parameters with weight decay ratio $\lambda$.



\subsection{Algorithm and Complexity analysis} \label{sec:IP}
Based on the above inferences,  we design the algorithm of \alg~as in Algorithm \ref{alg:new}. 
We will analyze the major computational complexity of the $l$-th layer of \alg\ in a training epoch. 
In the information propagating phase, for convenience, we first analyze the complexity of one connected pair $(v_{i}, v_{j})$. 
In the feature projection, node representation and feature representation are calculated by Eq. (\ref{eq:6}) and Eq. (\ref{eq:7}) with time complexity of $O(n)$ and $O(e)$ respectively. 
In the information propagation, the information representation associated with multiple nodes can be obtained through Eq. (\ref{eq:9}) with a time complexity of $O(e/2)$. 
In the information aggregation, the neighbourhood information between different nodes is aggregated through Eq. (\ref{eq:sum}) with a time complexity of $O(1)$. 
The time complexity obtained by combining the three phases is $O(n)$.

\begin{algorithm}[!t]
    \SetKwInOut{Input}{Input}
    \SetKwInOut{Output}{Output}
    \SetKwInOut{Init}{Init.}
    \SetKwInOut{Para}{Para.}
    \newcommand\mycommfont[1]{\ttfamily\textcolor{RoyalBlue}{#1}}
    \SetCommentSty{mycommfont}
    \Input{The undirected attributed network: $\mathcal{G}=\{\mathcal{V}, \mathcal{E}, \mathbf{F}, \phi, \psi \}$ }
    \Output{The node representation matrix $\mathbf{H}^{L}$ and the predicted matrix $\mathbf{Z}$}
    \Init{Node representation matrix $\mathbf{H}^{0} = \mathbf{F}$, global sharing learnable edge mapping matrix $\mathbf{W}^{r}$, and model parameters $\Theta$}
	\BlankLine
        \While{$\mathcal{L}$ does not converge}{ 
	\For {$l = 0, \ldots, L-1 $}{
        \tcc{Feature Projection}
        Compute node representation by Eq.(\ref{eq:6})\;
        Computer relation representation of node representation $\hat{\mathbf{r}}_{ij}^{l}$ of the feature bu Eq.(\ref{eq:7})\;
        
        \tcc{Information Propagation}
        Propagate the information $\mathbf{M}^{l}_{ij}$ from node $v_{i}$ to node $v_{j}$ by Eq.(\ref{eq:9})\;
        
        \tcc{Information Aggregation}
        Aggregate feature of neighbors to get node representation $\mathbf{H}^{l+1}$ by Eq.(\ref{eq:sum})
    }
    Computer the normalize representation b Eq.(\ref{eq:normalization})\;
    Compute the predicted matrix $\mathbf{Z}$ by Eq. (\ref{eq:z})\;
    Compute the model loss $\mathcal{L}$ by Eq. (\ref{eq:loss})\;
    Update model parameters $\Theta$ by the Adam optimizer\;
    }
    \caption{The \alg\ Algorithm}
    \label{alg:new}
\end{algorithm}

\section{{\bf Experiments}}
\label{sec:experiments}

%
In this section, we evaluate the effectiveness of \alg~on four network datasets by comparing with eight state-of-the-art methods. 
Specifically, we conduct node classification experiment to evaluate the end-to-end learning performance of our model, node clustering experiment to evaluate whether the learned node representations can fit the correct node class distribution.
Besides, for observing the intuitive display of node representation, the node visualization is conducted as an auxiliary experiment.
Finally, for giving a time Analysis and parameter sensitivity of our model, we conducted an explanatory ablation study on time complexity and sensitivity analysis of model parameters. We chose the ACM and DBLP datasets and visualized the model's efficiency and performance across various parameters.

The experiments are designed to answer the following research questions ({\bf RQs}):
\begin{itemize}[leftmargin=*]
	\item \textbf{RQ1.} {\em Does our \alg\ outperform the state-of-the-arts in Node Classification?}
    \item \textbf{RQ2.} {\em How does the clustering and visualisation of nodes perform?}
	\item \textbf{RQ3.} {\em How effective is each component of our model?}
	\item \textbf{RQ4.} {\em How is the time consumption and sensitivity to important parameters of our \alg?}
\end{itemize}
Section~\ref{sec:datasets} introduces the studied datasets.
Sections~\ref{sec:protocol} elaborates the experiment setup.
Results and findings are given in Sections \ref{sec:node}, \ref{sec:ablation} and \ref{sec:time}.

\subsection{Datasets} \label{sec:datasets}
Our assessments are compared against four sets of data, and their metrics are condensed in Table~\ref{tab:dataset}. 
In order to verify the wide applicability of our \alg\ algorithm, we handpicked the datasets from a variety of domains such as computer science, films, wikipedia, and others.
%
For node classification, a transductive approach is employed, wherein all edges are accessible during the training phase, and node labels are partitioned with 24\% for training, 6\% for validation, and 70\% for testing within each dataset.
%
They are detailed descriptions are listed as follows:
\begin{itemize}[leftmargin=*] 
    \item {\bf DBLP\footnote{\url{{http://web.cs.ucla.edu/~yzsun/data/}}}}: This dataset captures a computer science bibliography website, employing a prevalent subset across four domains with nodes symbolizing authors, papers, terms, and venues.
    \item {\bf ACM\footnote{\url{{https://libraries.acm.org/digital-library/acm-guide-to-computing-literature}}}}: This dataset concerns a movie-oriented website encompassing diverse genres like Action, Comedy, Drama, Romance, and Thriller.
    \item {\bf IMDB\footnote{\url{{https://www.kaggle.com/karrrimba/movie- metadatacsv}}}}: 
    This dataset also constitutes a citation network. We utilize the subset hosted within HAN, while retaining all edges, encompassing both paper citations and references.
    \item {\bf Freebase~}\cite{bollacker2008freebase}: 
    This dataset constitutes an extensive knowledge graph, extracted as a subgraph from 8 distinct genres: books, films, locations, music, individuals, sports, organizations, and businesses. The subgraph encompasses approximately 1,000,000 edges, following a method aligned with a prior survey \cite{yang2020heterogeneous}.
\end{itemize}


\begin{table}[!t]
	\centering
	\small
        \setlength{\tabcolsep}{6pt}
	\caption{Characteristics of the studied datasets.}
        \scalebox{.95}{
	\begin{tabularx}{0.5\textwidth}{p{1.3cm}|cXcXc}
		\toprule
		\midrule
		\emph{Node classification}    & \#Nodes.  & \#Node Types.  & \#Edges.  & \#Edge Types.  & \#Classes. \\
		\midrule
		DBLP      & \num{26128}   & \num{4}      & \num{239566}  & 6 & \num{4}  \\
		ACM       & \num{10942}   & \num{4}      & \num{547872}  & 6 & \num{3}  \\
		IMDB      & \num{21420}   & \num{4}      & \num{86642}   & 8 & \num{5}  \\
		Freebase  & \num{180098}  & \num{8}      & \num{1057688} & 36 & \num{7}  \\
            \bottomrule
	\end{tabularx}
        }
        \label{tab:dataset}
\end{table}

\subsection{Compared Methods} 
\label{sec:protocol}
\par\smallskip\noindent
%
We take eight graph learning competitors to evaluate the effectiveness and generalizability of our \alg\ approach in various settings.
In the below, we describe their key ideas in a high-level, and discuss why they are chosen to benchmark the experiments.

\begin{itemize}[leftmargin=*]
    \item {\bf GCN}~\cite{kipf2016semi}: It entailed a multi-layer graph convolutional neural network structure, wherein each convolutional layer specifically addresses first-order neighborhood attributes. By layering multiple convolutional tiers, it enables the potential for multi-order neighborhood information propagation to be realized.

    \item {\bf GAT}~\cite{velivckovic2017graph}: It was a graph neural network model based on an attention mechanism that assigns different weights to each node, and these weights affect the information transfer of the nodes in the graph.

    \item {\bf GTN}~\cite{yun2019graph}: It was a graph neural network model that leverages meta-path graphs to transform a heterogeneous graph and subsequently employs GCN for learning node embeddings. 
    
    \item {\bf HGT}~\cite{hu2020heterogeneous}: It was to combine the power of attention mechanisms and the Transformer architecture to handle and process complex heterogeneous graph data.

    \item {\bf HAN}~\cite{wang2019heterogeneous}: It was that different types of edges should have different weights, and different neighbouring nodes in the same type of edge have different weights, so it uses node level attention and semantic level attention.

    \item {\bf RGCN}~\cite{schlichtkrull2018modeling}: It considers the influence of different relations on nodes and proposes a method for fusing multiple relations in heterogeneous graphs, addressing the limitation that GCN does not consider node types.

    \item {\bf MAGNN}~\cite{fu2020magnn}: It captures structural and semantic information about the heterogeneous graph from neighbouring nodes and the meta-path contexts between them, and then performs inter-meta-path aggregation using an attention mechanism to fuse potential vectors obtained from multiple meta-paths into the final node embedding.

    \item {\bf Simple-HGN}~\cite{lv2021we}: It was based on the GAT and redesigned with three modules for learnable edge type embedding, residual connectivity, and $L_{2}$ normalization of the output embedding.

    \item {\bf SeHGNN}~\cite{yang2023simple}: It aimed to capture structural insights by pre-computing neighbor aggregation through a lightweight mean aggregator. Furthermore, it expands the receptive scope by implementing a single-layer structure that incorporates extended metapaths.

    \item {\bf MH-GCN}~\cite{li2022multi}: It aims to tackle the challenge of capturing heterogeneous structural signals in multiplex networks and demonstrates significant superiority in various network analysis tasks.

\end{itemize}

\subsection{Evaluation Protocol}
To perform a fair comparison, the experiments are benchmarked in heterogeneous graph settings.

Micro-F1 and Macro-F1 are metrics used to evaluate classifier performance in multi-category classification problems.
%
Micro-F1 is calculated as the harmonic mean of precision and recall over the entire dataset, namely, 
\begin{equation} \label{eq:micro}
    \textnormal{F}_{micro} =\frac{k \times P_{\text {sum }} \times R_{\text {sum }}}{k^2 \times P_{\text {sum }}+R_{\text {sum }}},
\end{equation}
where $P_{\text{sum}}$ and $R_{\text{sum}}$ represent the sum of precision and recall across all categories in the dataset, respectively, while $k$ denotes the total number of categories.

It is consistent with a similar sample size for each category.  
%
Macro-F1 prioritizes the calculation of precision and recall for each category first, and then takes the average, namely, 
\begin{equation}\label{eq:macro}
    \textnormal{F}_{macro} =\frac{k \times P_{\text {ave}} \times R_{\text {ave}}}{k^2 \times P_{\text {ave }}+R_{\text {ave}}},
\end{equation}
where $P_{\text{ave}}$ and $ R_{\text{ave}} $ denote the average precision and recall per category, respectively.
The evaluation of node classification performance employs the Adjusted Rand index (ARI). 
Additionally, the assessment of node clustering performance incorporates both the Adjusted Rand index (ARI) and the Normalized Mutual Information (NMI). 
The computation of the Adjusted Rand index (ARI) is outlined as follows:
\begin{equation}
\textnormal{ARI}=\frac{\textnormal{RI}-E[\textnormal{RI}]}{\max (\textnormal{RI})-E[\textnormal{RI}]},
\end{equation}
\begin{equation}
    \textnormal{RI} =\frac{U + V}{C_2^{n_{\text {samples}}}},
\end{equation}
where $\textnormal{RI}$ is a metric used to assess the similarity between two clustering results, $U$ and $V$ denote the number of pairs of samples that are either in the same category or in different categories in the ground-truth and clustering result, respectively.

\begin{table*}[!t]
	\centering
	\caption{ Node classification performance evaluation. "OOM" denotes instances where the models exhausted memory resources. Higher values are indicative of better performance. The most optimal outcomes are highlighted in \textbf{bold}. \emph{w/o} FGL: no edge-to-node mapping or dot product; \emph{w/o} $L_{2}$: no $L_{2}$ normalization for node mapping; \emph{w/o} non-linear encoding (NLE): only linear transformation for nodes; \emph{w/o} edge information (EI): random edge initialization instead of type-based encoding.}
	\setlength{\tabcolsep}{4pt}
        \scalebox{.95}{
	\begin{tabular}{c|cc|cc|cc|cc}
		\toprule
		\midrule
		   & \multicolumn {2}{c|}{DBLP}
		   & \multicolumn {2}{c|}{ACM}
              & \multicolumn {2}{c|}{IMDB}
              & \multicolumn {2}{c}{Freebase}\\
		
		& micro-f1   & macro-f1   & micro-f1     & macro-f1 
            & micro-f1   & macro-f1   & micro-f1     & macro-f1\\
		\midrule
         GCN       & $90.33 \pm 0.36$    
            & $89.68 \pm 0.43$    
            & $92.28 \pm 0.41$
		  & $92.30 \pm 0.43$                
            & $63.81 \pm 0.74$  
            & $56.32 \pm 1.51$
            & $59.51 \pm 0.34$
            & $31.01 \pm 0.66$
		\\
  GAT       & $92.26 \pm 0.64$    
            & $91.66 \pm 0.65$    
            & $91.01 \pm 1.28$
		  & $91.03 \pm 1.26$                
            & $65.32 \pm 1.08$  
            & $59.53 \pm 2.26$
            & $62.37 \pm 0.42$
            & $40.85 \pm 0.72$
		\\
    GTN     & $93.97 \pm 0.54$    
            & $93.52 \pm 0.55$    
            & $91.20 \pm 0.71$
		  & $91.31 \pm 0.70$                
            & $65.14 \pm 0.45$  
            & $60.47 \pm 0.98$
            & OOM
            & OOM
		\\
    HGT     & $93.49 \pm 0.25$    
            & $93.01 \pm 0.23$    
            & $91.00 \pm 0.76$
		  & $91.12 \pm 0.76$                
            & $67.20 \pm 0.57$  
            & $63.00 \pm 1.19$
            & $60.51 \pm 1.16$
            & $29.28 \pm 2.52$
		\\
		HAN & $93.57 \pm 0.33$    
            & $93.09 \pm 0.34$    
            & $88.11 \pm 0.93$
		  & $88.06 \pm 1.02$                
            & $65.06 \pm 1.20$  
            & $58.54 \pm 2.39$
            & $56.26 \pm 0.87$
            & $21.66 \pm 2.29$
		\\
        RGCN & $92.15 \pm 1.22$    
            & $91.66 \pm 1.30$    
            & $90.54 \pm 0.89$
		  & $90.65 \pm 0.86$                
            & $61.73 \pm 0.94$  
            & $51.37 \pm 2.35$
            & $58.46 \pm 1.46$
            & $44.11 \pm 1.05$
		\\
  MAGNN     & $93.35 \pm 0.59$    
            & $92.85 \pm 0.66$    
            & $91.44 \pm 0.67$
		  & $91.50 \pm 0.68$                
            & $65.32 \pm 0.66$  
            & $57.22 \pm 1.59$
            & OOM
            & OOM
		\\
  
  MH-GCN    & $94.18 \pm 0.47$    
            & $93.84 \pm 0.48$    
            & OOM
		  & OOM
            & $60.11 \pm 1.32$  
            & $58.57 \pm 1.27$
            & OOM
            & OOM
		\\
    Simple-HGN
            & $94.10 \pm 0.41$    
            & $93.65 \pm 0.42$    
            & $93.31 \pm 0.36$
		    & $93.40 \pm 0.36$                
            & $67.04 \pm 0.38$  
            & $64.14 \pm 0.49$
            & $\underline{65.52 \pm 0.55}$
            & $45.26 \pm 2.24$

		\\
  SeHGNN    & $\underline{95.42 \pm 0.17}$    
            & $\underline{95.06 \pm 0.17}$    
            & $\underline{93.98 \pm 0.36}$
		    & $\underline{94.05 \pm 0.35}$                
            & $\underline{69.17 \pm 0.43}$  
            & $\underline{67.11\pm0.25}$
            & $65.08 \pm 0.66$
            & $\underline{51.87 \pm 0.86}$
		\\
  \midrule
  \emph{w/o} FGL    
            & $92.82\pm0.69$ 
            & $92.30\pm0.73$  
            & $92.51\pm0.21$
		    & $92.55\pm0.23$                
            & $67.47\pm0.54$  
            & $61.13\pm1.77$
            & $68.32\pm0.59$
            & $51.74\pm0.74$
		\\
    \emph{w/o} $L_{2}$    
            & $95.52\pm0.35$   
            & $95.18\pm0.38$    
            & $93.08\pm0.48$
		    & $93.10\pm0.47$                
            & $68.32\pm0.58$  
            & $63.13\pm1.15$
            & $67.19\pm0.61$
            & $47.03\pm2.49$
		\\
    \emph{w/o} NLE    
            & $95.56\pm0.12$
            & $95.20\pm0.11$
            & $93.83\pm0.29$
		    & $93.88\pm0.30$                
            & $68.06\pm0.42$  
            & $62.25\pm0.09$
            & $68.49\pm0.15$
            & $52.05\pm0.85$
		\\
    \emph{w/o} EI   
            & $92.63\pm0.48$
            & $92.12\pm0.52$
            & $91.82\pm0.76$
		    & $91.83\pm0.79$                
            & $65.31\pm1.13$  
            & $58.41\pm1.85$
            & $68.02\pm0.26$
            & $51.11\pm0.99$
		\\
  \midrule
    \alg      & $\mathbf{95.98\pm0.01} $
            & $\mathbf{95.68\pm0.01} $    
            & $\mathbf{94.42\pm0.01} $
		    & $\mathbf{94.47\pm0.01} $  
            & $\mathbf{69.36\pm0.04} $  
            & $\mathbf{67.29\pm0.02} $
            & $\mathbf{69.04\pm0.03} $
            & $\mathbf{53.18\pm0.09} $ \\
        \midrule
    \emph{p}-value
            & 0.012
            & 0.014
            & 0.018
            & 0.020
            & 0.025
            & 0.028
            & 0.015
            & 0.022 \\
    \emph{Improve}(\%)     
            & 0.59
            & 0.65
            & 0.47
		    & 0.45
            & 0.27
            & 0.27
            & 5.38
            & 2.53 \\
		\bottomrule
	\end{tabular}
        }
	\label{tab:algorithm}
\end{table*}

$C_i^t$ is the ground-truth label of each node, and $C_i^p$ is the label predicted by node embedding. And Normalized Mutual Information (NMI) is computed by:
\begin{equation}
\textnormal{NMI}\left(C_i^t, C_i^p\right)=\frac{\textnormal{I}\left(C_i^t, C_i^p\right)}{\left[\textnormal{H}\left(C_i^t\right)+\textnormal{H}\left(C_i^p\right)\right] / 2},
\end{equation}
where $\textnormal{I}(\cdot)$ and $\textnormal{H}(\cdot)$ denote the mutual information and entropy functions. 
The value of NMI ranges from 0 to 1, where 1 indicates perfect clustering results, and 0 signifies that the clustering outcome is indistinguishable from random assignment.
Their calculations are outlined as follows:
\begin{equation}
\textnormal{I}\left(C_i^t, C_i^p\right)=\sum_i \sum_j \frac{\left|C_i^t \cap C_i^p\right|}{n} \log \frac{N \cdot\left|C_i^t \cap C_i^p\right|}{\left|C_i^t\right|\left|C_i^p\right|} \\,
\end{equation}
\begin{equation}
    \textnormal{H}(C)=-\sum_i \frac{C_i}{N} \log \frac{C_i}{N}.
\end{equation}

Both of ACC and NMI can represent the correlation between the predicted label and the ground-truth label. 
And the higher value of them indicates the better performance of node clustering.
All the experiments are run on a computer server that has a 2.30 GHz Intel GlodXeon GD 5218 with 16 cores, and Quadro RTX6000 24GB * 1.

\subsection{Node Classification (\textbf{RQ1})} \label{sec:node}
We compare the performance of our model on the node classification assignment to that of current best practices. The outcomes are presented in Tables \ref{tab:algorithm} and \ref{tab:algorithm2}, with the finest highlighted in bold.The baseline includes homogeneous graph methods, heterogeneous graph methods, and graph representation learning methods.

As we can see, \alg\  exhibits outstanding efficiency on all tested networks. 
Regarding micro-F1 and macro-F1, our \alg\ outperforms the state-of-the-art HGNN model SeHGNN by an average of 1.51\% and 0.64\%, respectively, across all datasets. 
Given that the performance gain in node classification tasks described in some recent publications is typically around 0.5-1.0\%, our \alg's performance boost is noteworthy. 
Furthermore, we find that \alg\ outperforms rival approaches on large heterogeneous networks with multi-typed nodes (\emph{e.g.}, Freebase), attaining 5.3\% and 0.91\% improvement in micro-F1 and macro-F1 on the Freebase network, respectively. 
%
The potential explanation lies in the effectiveness of our \alg\ in acquiring node representations tailored for classification through the exploration of interactions among various relations carrying distinct levels of significance (\emph{i.e.}, weights). 
This aspect is often disregarded by heterogeneous network embedding techniques centered on scalar representation learning.


\subsection{Node Clustering and Visualization (\textbf{RQ2})} \label{sec:link}
\subsubsection{Node Clustering}
In the clustering task, the output of the penultimate layer is preserved, which will be fed into the downstream clustering model. 
We set the number of clusters according to the number of labels, and the k-means algorithm is used as the downstream clustering method. 
For reducing the impact of the ground-truth information on the unsupervised learning task, we only use 3\% labels of each dataset to train the full-supervised model.

\begin{table}[!t]
	\centering
	\small
	\caption{Comparison Results in terms of Precision (\%) and Mean Average Precision (MAP) (\%) in the node clustering task on the four datasets. The most optimal results are highlighted in bold and the second optimal results is \underline{underlined}. "OOM" denotes instances where the models exhausted memory resources.} 
	\setlength{\tabcolsep}{4pt}
    \scalebox{1}{
	\begin{tabular}{c|cc|cc}
		\toprule
		\midrule
		   & \multicolumn {2}{c|}{DBLP}
		   & \multicolumn {2}{c}{ACM}\\
              
		& Precision(\%)    & MAP(\%)    & Precision(\%)    & MAP(\%)\\
		\midrule
    GCN     &   90.6338
            &   84.4697
            &   92.7289
		    &   88.5327
		\\
    GAT     &   93.3450
            &   88.8479
            &   92.4457
		    &   88.1196
		\\
    HAN     &  93.7676
            &  89.5615
            &  89.2823
		    &  83.4253
		\\
    RGCN    &  92.3239
            &  87.1350
            &  89.3295
		    &  83.8275
		\\
    MAGNN   &  93.1690
            &  88.4674
            &  91.9735
		    &  87.4646
		\\
    MH-GCN  &  94.0845
            &  90.0175
            &  OOM
		    &  OOM
		\\
 Simple-HGN &  94.7887
            &  91.2015
            &  93.2483
		    &  89.3293
		\\
  SeHGNN    &  \underline{95.6338}
            &  \underline{92.7124}
            &  \underline{94.5231}
		    &  \underline{91.2279}
		\\
  \midrule
  \alg      &   $\mathbf{95.9485}$
            &   $\mathbf{93.0509}$
            &   $\mathbf{94.5103}$
	        &   $\mathbf{94.4287}$
            \\
		\bottomrule
    & \multicolumn {2}{c|}{IMDB}
    & \multicolumn {2}{c}{Freebase}\\
              
    & Precision(\%)    & MAP(\%)    & Precision(\%)    & MAP(\%)\\
		\midrule
    GCN     &  45.5028
            &  55.4675
            &  59.4827
		    &  44.3824
		\\
    GAT     &  48.5634
            &  57.3882
            &  62.8592
		    &  49.2971
		\\
    HAN     &  78.3135
            &  57.0063
            &  57.4712
	        &  39.9496  
		\\
    RGCN    &  \underline{76.5646}
            &  54.8855
            &  58.6566
		    &  46.0053
		\\
    MAGNN   &  47.5015
            &  56.8387
            &  OOM
		    &  OOM
		\\
    MH-GCN  &  61.1570
            &  51.5787
            &  OOM
		    &  OOM
		\\
 Simple-HGN &  50.4060
            &  58.0322
            &  \underline{65.8944}
		    &  \underline{51.3584}
		\\
  SeHGNN    &  75.1530
            &  \underline{58.9580}
            &  64.1882
		    &  49.6740
		\\
  \midrule
  \alg      &   \textbf{78.2957}
            &   \textbf{59.4462}
            &   \textbf{68.2298}
		    &   \textbf{55.3684}    
            \\
		\bottomrule
	\end{tabular}
        }
	\label{tab:algorithm2}
\end{table}

As shown in Table \ref{tab:nodeclustering}, Models based on meta-path or graph attention mechanisms perform relatively poorly on DBLP, ACM and Freebase datasets, especially with respect to NMI values.
In attention-based models, the interactions between nodes are achieved through attention weights. However, excessive reliance on neighboring node information may lead to issues of information leakage and over-smoothing.
Path-based models may encounter two challenges in node clustering. 
Firstly, inappropriate selection of meta-paths can result in insufficient reflection of relationships between nodes, thus impacting the accuracy and expressive power of node aggregation. 
Secondly, different meta-paths may encompass diverse association information, leading to difficulties in effectively integrating their contributions during aggregation.

\begin{table*}[!t]
	\centering
	\caption{Comparison Results in trems of ARI(\%) and NMI(\%) in the {\bf Node Clustering} task on the four datasets. The best result in trems of each evaluation measure is highlighted in bold.} 
	\setlength{\tabcolsep}{4pt}
        \scalebox{1}{
    	\begin{tabular}{c|cc|cc|cc|cc}
    		\toprule
    		\midrule
    		& \multicolumn {2}{c|}{DBLP}
    		  & \multicolumn {2}{c|}{ACM}
                & \multicolumn {2}{c|}{IMDB}
                & \multicolumn {2}{c}{Freebase} \\
                
    		& ARI(\%)    & NMI(\%)    & ARI(\%)    & NMI(\%)
                & ARI(\%)    & NMI(\%)    & ARI(\%)    & NMI(\%) \\
    		\midrule
        GCN     & $61.15 \pm 2.67$  
                & $55.55 \pm 2.05$    
                & $65.59 \pm 1.20$ 
                & $62.33 \pm 0.82$
    		    & $5.83 \pm 0.84$          
                & $8.55 \pm 0.44$  
                & $4.10 \pm 1.01$
                & $10.15 \pm 0.76$
    		\\
        GAT     & $53.70 \pm 3.54$
                & $52.26 \pm 2.46$    
                & $34.33 \pm 3.40$
                & $44.87 \pm 1.76$
                & $8.54 \pm 0.80$               
                & $11.09 \pm 0.81$  
                & $9.75 \pm 1.78$
                & $18.94 \pm 2.20$
    		\\
        HGT     & $74.86 \pm 1.48$
                & $69.19 \pm 1.77$    
                & $67.34 \pm 5.99$     
                & $63.93 \pm 3.53$
                & $3.20 \pm 2.33$              
                & $4.17 \pm 3.04$  
                & $4.52 \pm 2.80$
                & $8.06 \pm 1.12$
    		\\
        HAN     &  $76.71 \pm 1.25$
                &  $70.12 \pm 1.14$
                &  $43.50 \pm 1.28$
    		    &  $48.09 \pm 2.36$
                &  $4.18  \pm 0.34$
                &  $5.60  \pm 0.39$
                &  $4.02  \pm 1.09$
                &  $7.03  \pm 0.29$
    		\\
        RGCN    &  $0.28 \pm 0.18$
                &  $0.50 \pm 0.09$
                &  $23.17 \pm 0.74$
    		    &  $30.65 \pm 3.38$
                &  $0.20 \pm 0.00$
                &  $0.52 \pm 0.00$
                &  $5.77 \pm 1.24$
                &  $6.13 \pm 0.07$
    		\\
      MAGNN     &  $\underline{83.47 \pm 1.57}$
                &  $\underline{77.51 \pm 1.87}$
                &  $65.71 \pm 4.12$
                &  $62.13 \pm 4.16$
    		    &  $7.79 \pm 1.59$
                &  $9.13 \pm 0.84$
                &  OOM
                &  OOM
    		\\
      MH-GCN    &  $0.31 \pm 0.54$
                &  $2.42 \pm 1.17$
                &  OOM
                &  OOM
    		    &  $\underline{10.16 \pm 8.60}$
                &  $9.37 \pm 7.68$
                &  OOM
                &  OOM
    		\\
      Simple-HGN     
                &  $67.13 \pm 9.92$
                &  $64.32 \pm 5.19$
                &  $74.57 \pm 2.17$
                &  $70.20 \pm 1.82$
    		    &  $8.72 \pm 0.53$
                &  $\underline{11.42 \pm 0.57}$
                &  $13.30 \pm 1.64$
                &  $\underline{21.05 \pm 2.48}$
    		\\
      SeHGNN    &   $82.61 \pm 0.88$
                &   $77.20 \pm 0.75$
                &   $\underline{77.29 \pm 1.45}$
    		    &   $\underline{71.87 \pm 1.58}$
                &   $8.33  \pm 0.57$
                &   $10.31 \pm 0.72$
                &   $\underline{13.70 \pm 0.95}$
                &   $17.58 \pm 0.52$
    		\\
      \midrule
      \alg      & $\mathbf{84.46\pm1.36} $
                & $\mathbf{79.73\pm1.41} $
                & $\mathbf{77.70\pm0.07} $    
                & $\mathbf{72.42\pm0.88} $
    		    & $\mathbf{10.80\pm1.34} $
                & $\mathbf{12.39\pm0.77} $  
                & $\mathbf{14.81\pm0.84} $
                & $\mathbf{22.71\pm1.67} $ \\
        \midrule
        \emph{p}-value
            & 0.014
            & 0.018
            & 0.020
            & 0.022
            & 0.015
            & 0.018
            & 0.012
            & 0.016 \\
        \emph{Improve}(\%)     
            & 1.19
            & 2.86
            & 0.53
		    & 0.77
            & 6.30
            & 8.49
            & 8.11
            & 7.88 \\
    		\bottomrule
    	\end{tabular}
        }
	\label{tab:nodeclustering}
\end{table*}

\begin{figure*}[!t]
    \centering
    \begin{flushleft}
	\begin{subfigure}[t]{0.16\linewidth}
		\includegraphics[width=\textwidth]{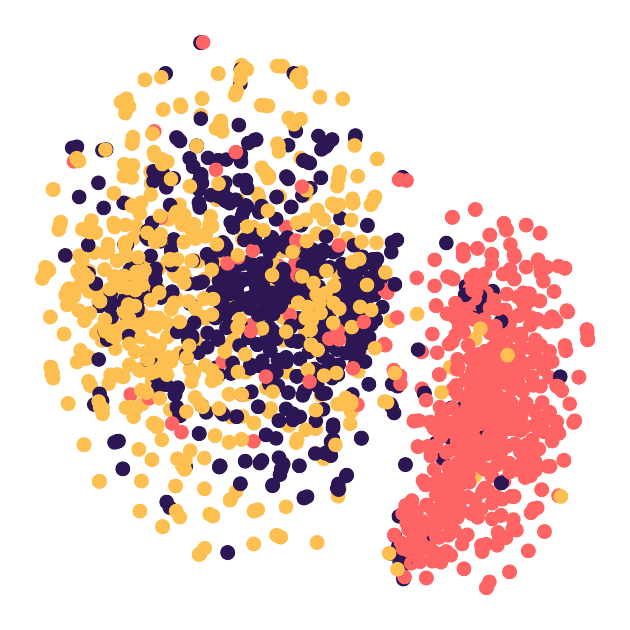}
		\caption{HAN}
	\end{subfigure}
         \begin{subfigure}[t]{0.16\linewidth}
		\includegraphics[width=\textwidth]{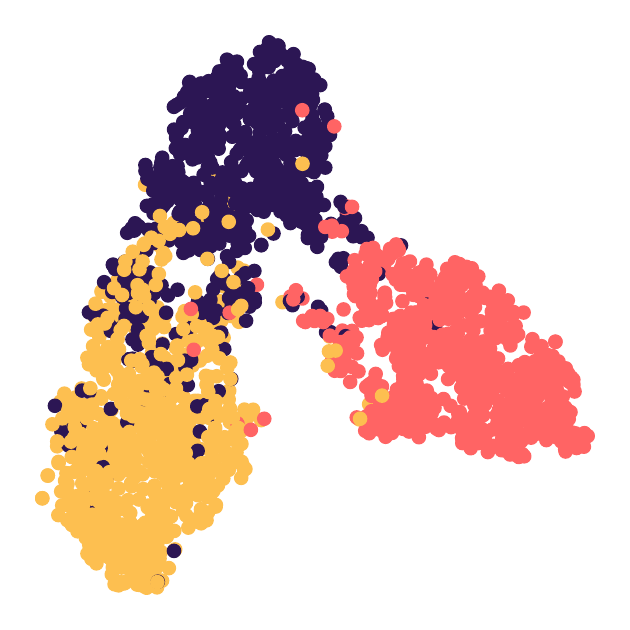}
		\caption{Simple-HGN}
	\end{subfigure}
	\begin{subfigure}[t]{0.16\linewidth}
		\includegraphics[width=\textwidth]{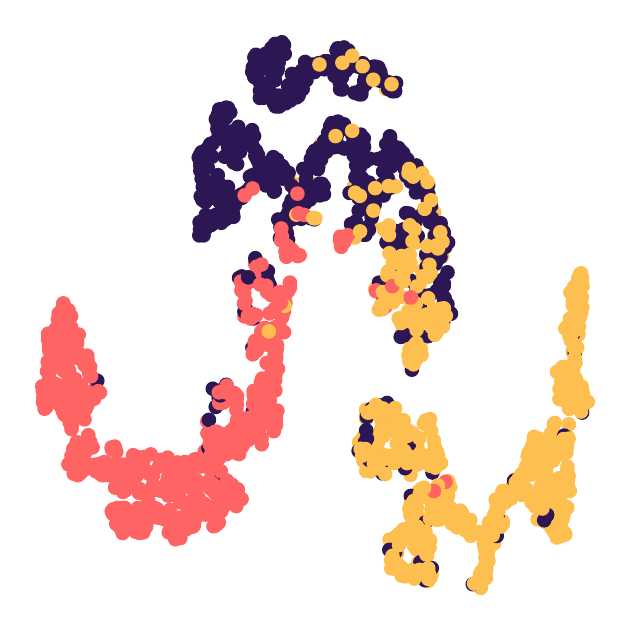}
		\caption{HGT}
	\end{subfigure}
        \begin{subfigure}[t]{0.16\linewidth}
		\includegraphics[width=\textwidth]{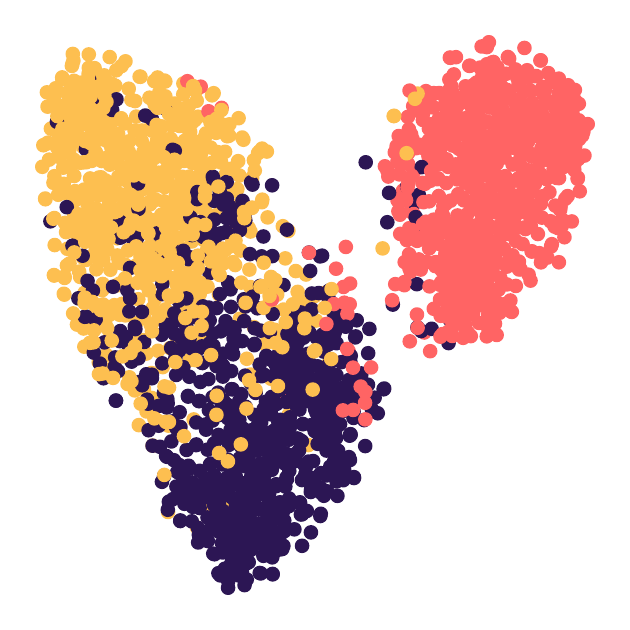}
		\caption{MAGNN}
	\end{subfigure}
        \begin{subfigure}[t]{0.16\linewidth}
		\includegraphics[width=\textwidth]{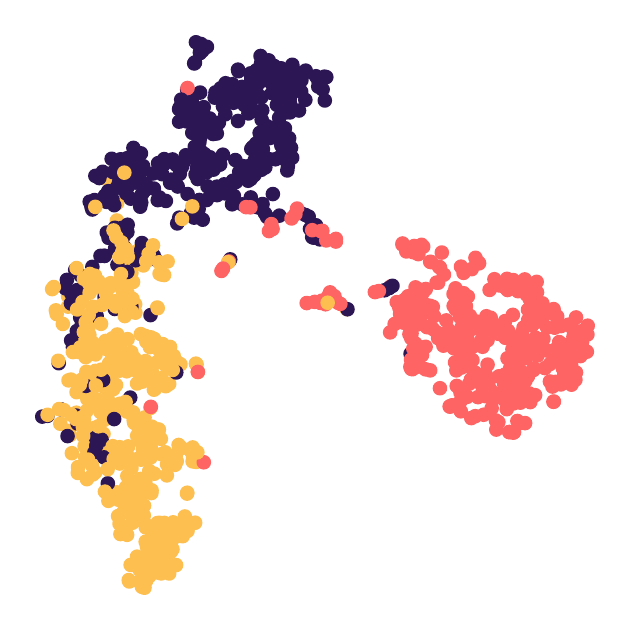}
		\caption{SeHGNN}
	\end{subfigure}
        \begin{subfigure}[t]{0.16\linewidth}
		\includegraphics[width=\textwidth]{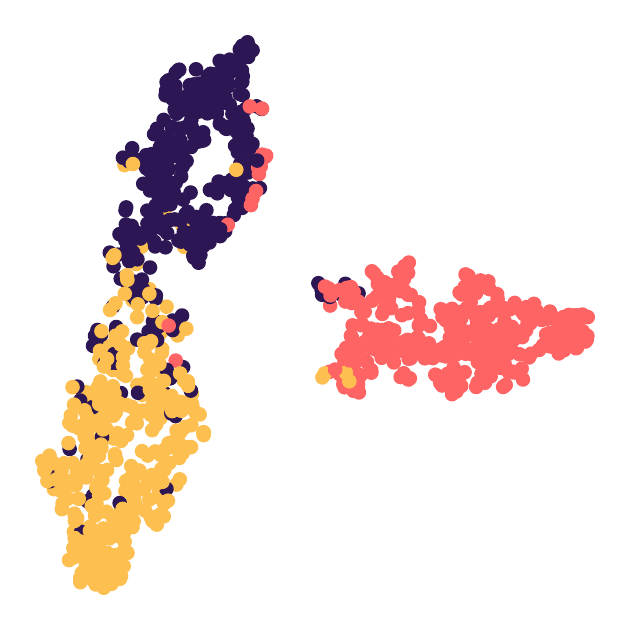}
		\caption{\alg}
	\end{subfigure}
	\caption{Visualization of embedding on ACM. Nodes with different labels are differentiated by colors.}
	\label{fig:ablation}
    \end{flushleft}
\end{figure*}

\begin{figure*}[!t]
    \centering
    \begin{flushleft}
	\begin{subfigure}[t]{0.16\linewidth}
		\includegraphics[width=\textwidth]{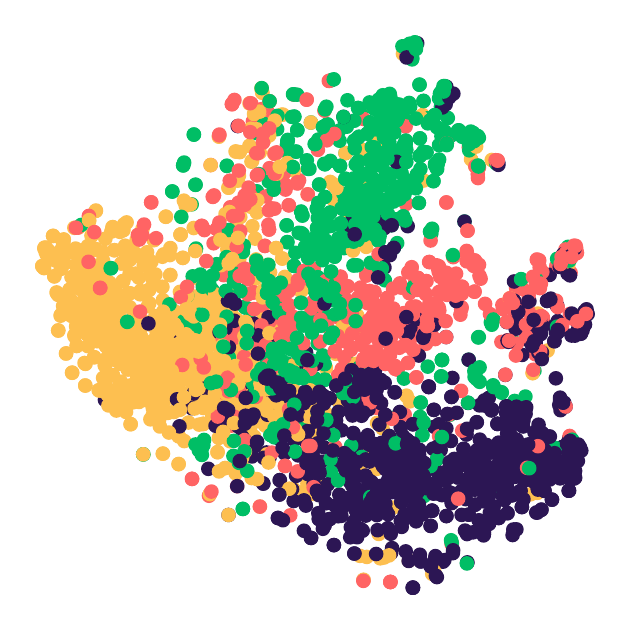}
		\caption{HAN}
	\end{subfigure}
         \begin{subfigure}[t]{0.16\linewidth}
		\includegraphics[width=\textwidth]{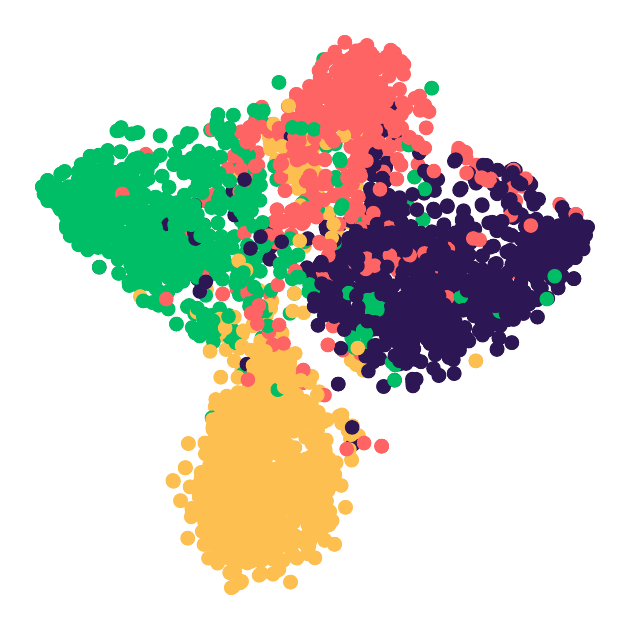}
		\caption{Simple-HGN}
	\end{subfigure}
	\begin{subfigure}[t]{0.16\linewidth}
		\includegraphics[width=\textwidth]{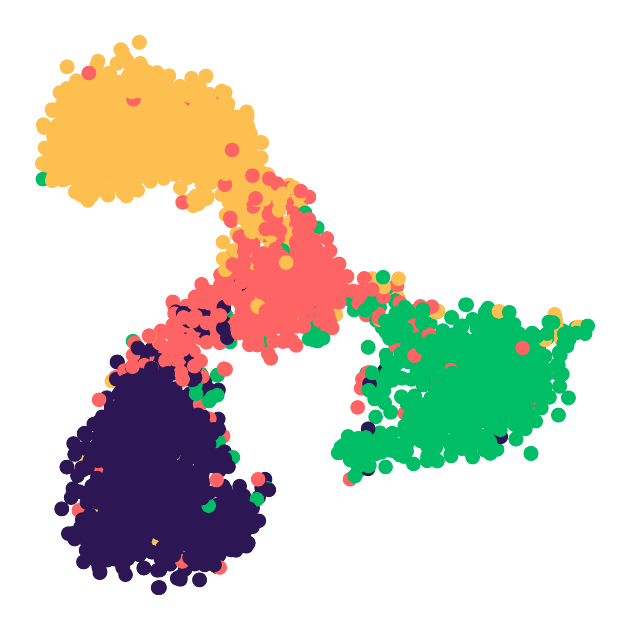}
		\caption{HGT}
	\end{subfigure}
        \begin{subfigure}[t]{0.16\linewidth}
		\includegraphics[width=\textwidth]{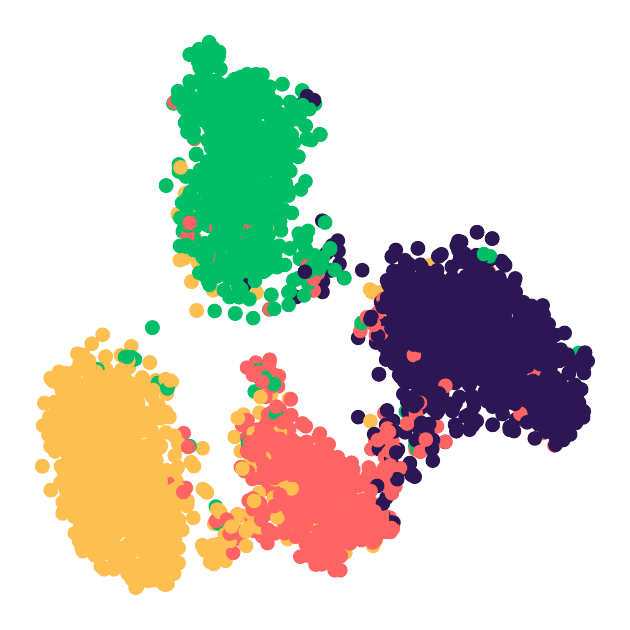}
		\caption{MAGNN}
	\end{subfigure}
        \begin{subfigure}[t]{0.16\linewidth}
		\includegraphics[width=\textwidth]{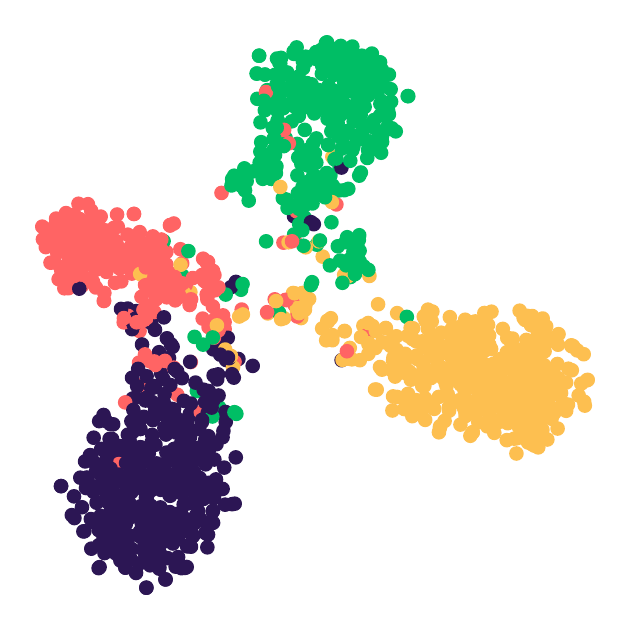}
		\caption{SeHGNN}
	\end{subfigure}
        \begin{subfigure}[t]{0.16\linewidth}
		\includegraphics[width=\textwidth]{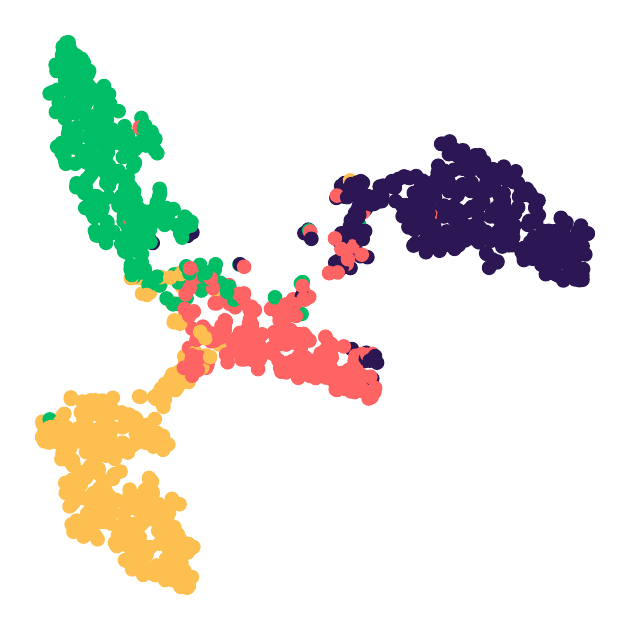}
		\caption{\alg}
	\end{subfigure}
	\caption{Visualization of embedding on DBLP. Nodes with different labels are differentiated by colors.}
	\label{fig:ablation2}
    \end{flushleft}
\end{figure*}

\subsubsection{Visualization} 
For a more intuitive comparison, we adopt T-SNE \cite{liao2018attributed} to obtain 2-dimensional node representations, and each node will be colored differently according to the ground-truth label.
For limited space, we choose to display the visualization results of ACM and DBLP instead of all datasets.
From Figs. \ref{fig:ablation} and \ref{fig:ablation2}, we can find \alg~provides the best visualization performance. 
The visualization result of HAN is a little bit messy. 
In the method, attention mechanisms are used to determine the importance of nodes and meta-paths. 
However, at higher levels, the attention weights may become more averaged, blurring the differences between nodes. 

\begin{figure}[!h]
    \centering
    \begin{flushleft}
	\begin{subfigure}[t]{0.95\linewidth}
		\includegraphics[width=\textwidth]{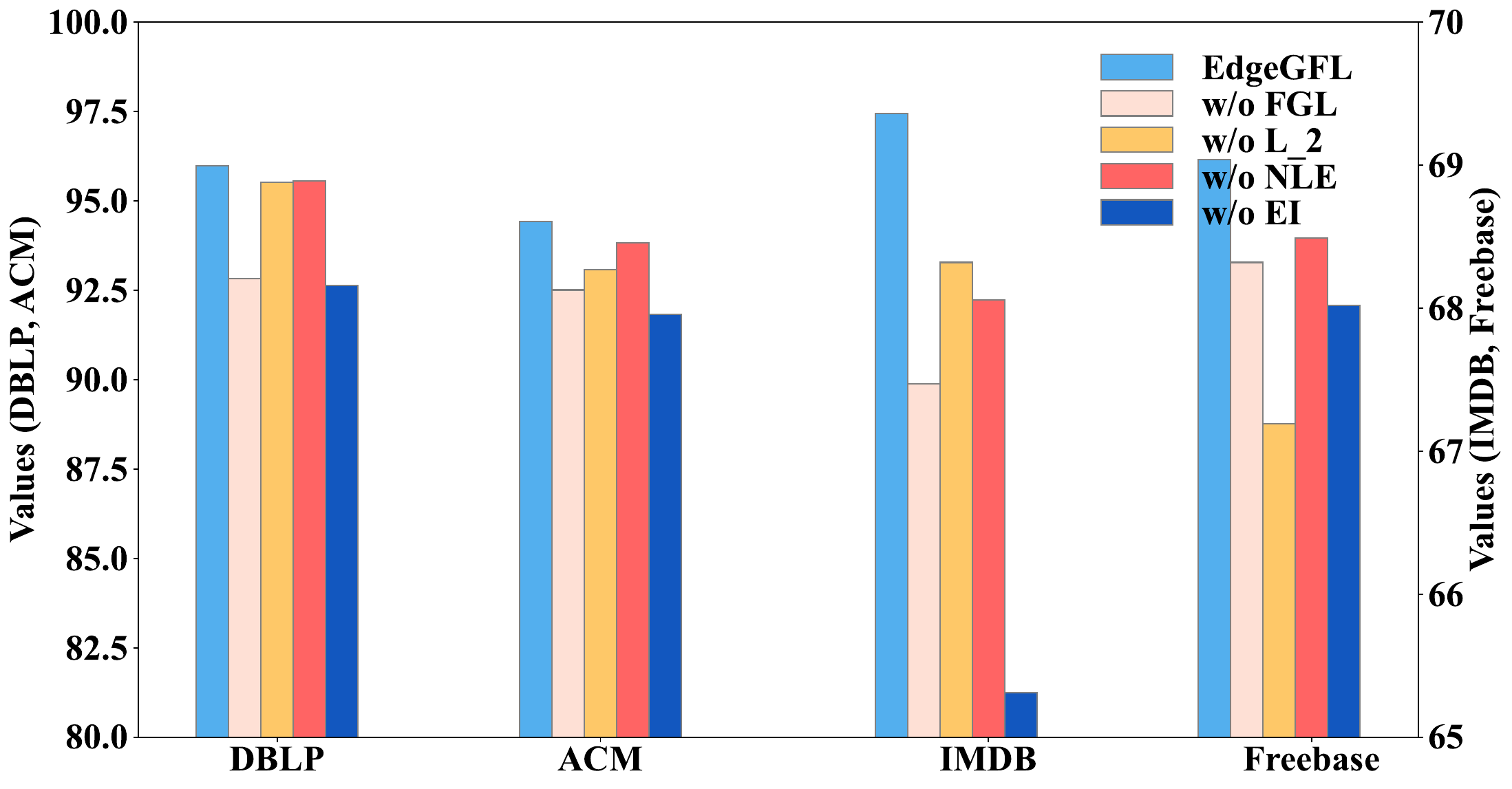}
		\caption{Micro-F1 score}
	\end{subfigure}
	\begin{subfigure}[t]{0.95\linewidth}
		\includegraphics[width=\textwidth]{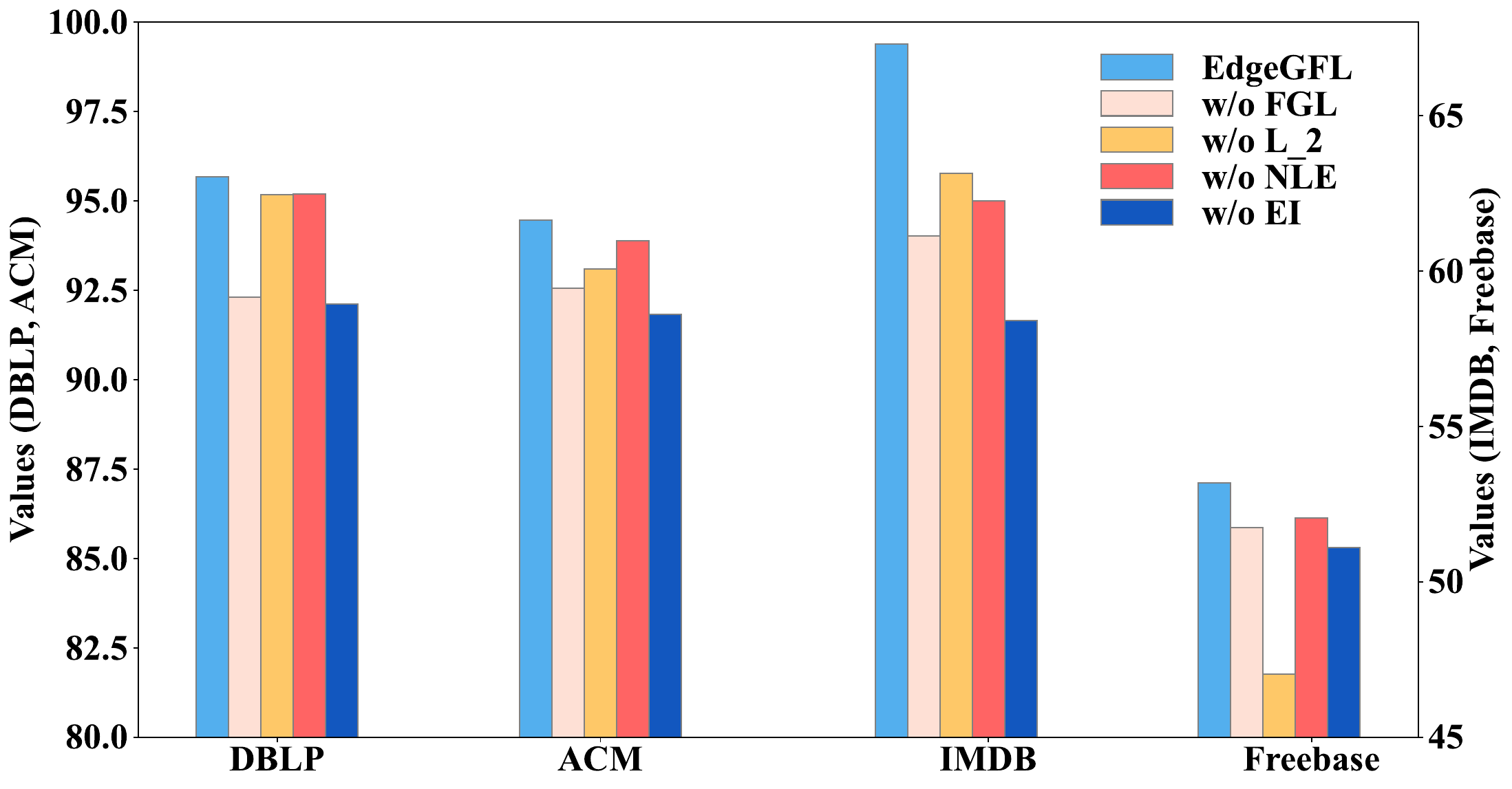}
		\caption{Macro-F1 score}
	\end{subfigure}
	\caption{Experimental results of ablation study}
	\label{fig:ablation33}
    \end{flushleft}
\end{figure}

As a result, the distribution and clustering of nodes in the visualization are affected, leading to less accurate node representations.
Simple-HGN is a graph-based model that generates node embeddings in the penultimate layer, which may suffer from poor visualization due to high dimensionality, insufficient feature richness, and limited learning capacity in capturing complex relationships. 
Addressing these issues through dimensionality reduction, feature enrichment, and model enhancement can improve the visualization quality.
As we can see, for HGT, MAGNN and SeHGNN, node representations with different labels are more distinct from each other, while those having the same label are more compact, which to a large extent demonstrates that the attention-based or mate-path-based models can integrate the network structure and node attribute information more rationally.
Specifically, \alg~can learn the best node distribution, which can distinctly separate nodes with different labels and thus provide convenience for both node classification and node clustering tasks.

\begin{figure*}[!h]
    \centering
    \begin{flushleft}
	\begin{subfigure}[t]{0.32\linewidth}
		\includegraphics[width=\textwidth]{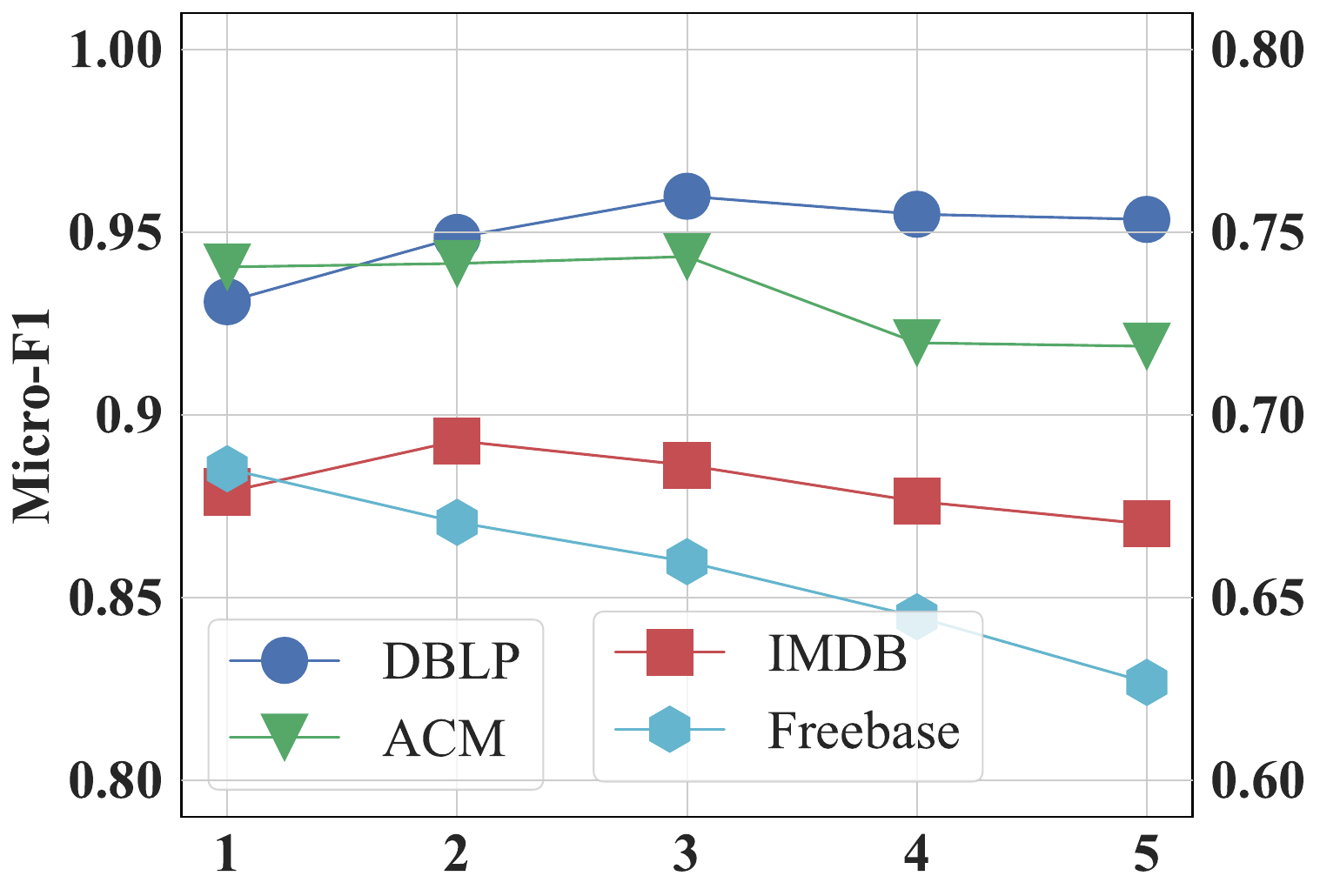}
		\caption{Micro-F1 score \emph{w.r.t.} \#layers}
		\label{fig:layers}
	\end{subfigure}
	\begin{subfigure}[t]{0.32\linewidth}
		\includegraphics[width=\textwidth]{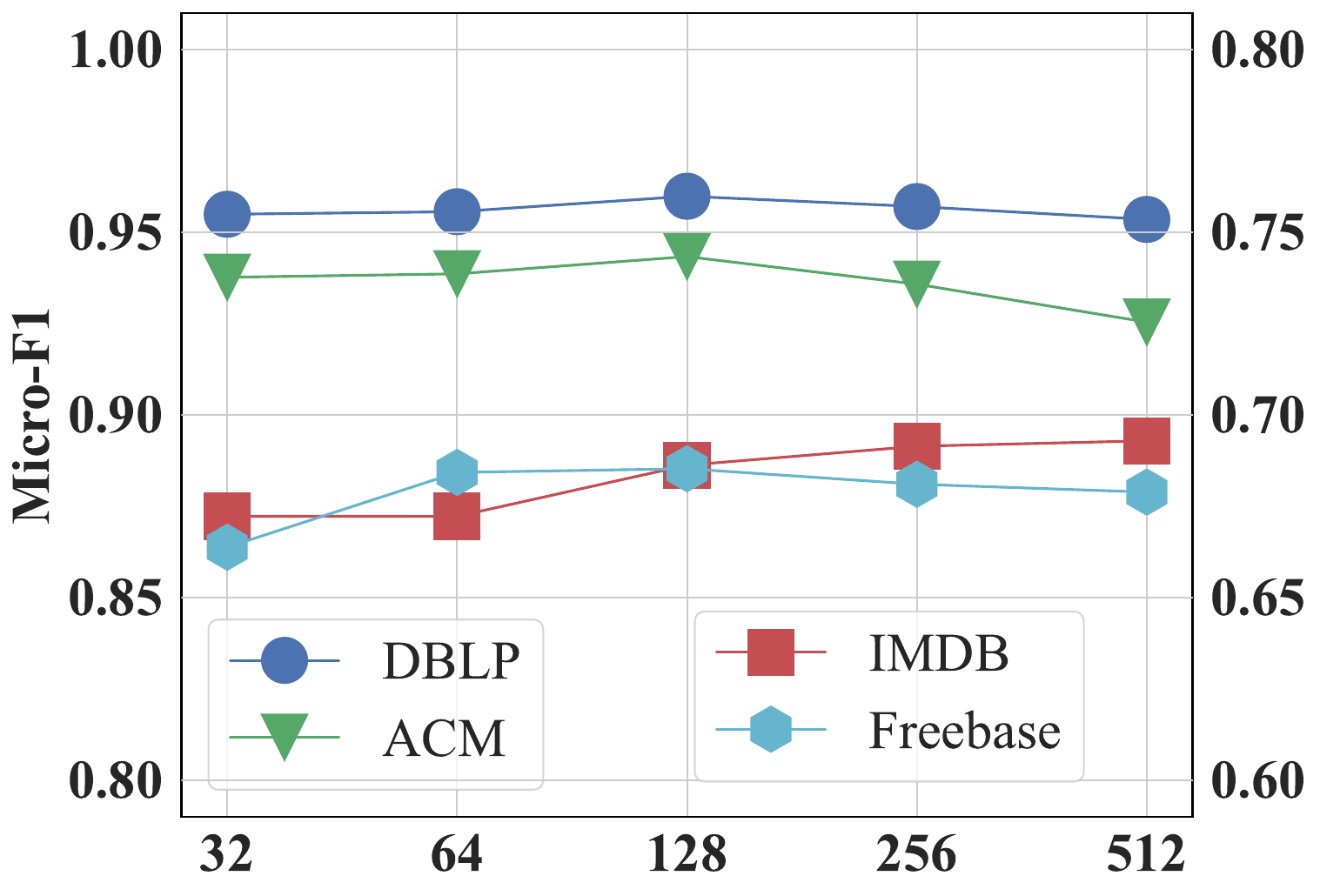}
		\caption{Micro-F1 score \emph{w.r.t.} \#dimensions}
		\label{fig:dim}
	\end{subfigure}
	\begin{subfigure}[t]{0.32\linewidth}
		\includegraphics[width=\textwidth]{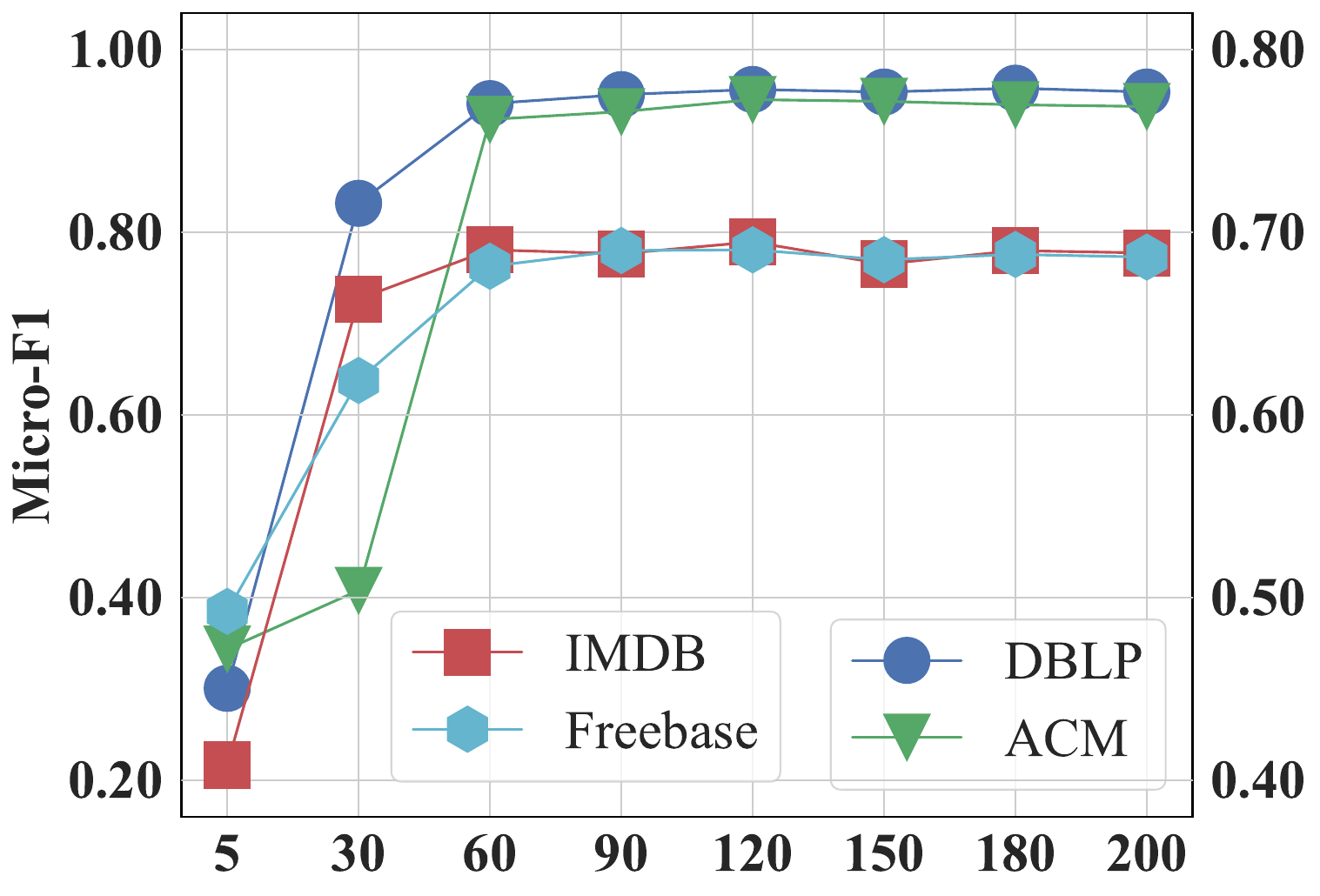}
		\caption{Micro-F1 score \emph{w.r.t.} \#rounds}
		\label{fig:rounds}
	\end{subfigure}
	\caption{
            Micro-F1 outcomes depicting the parameter sensitivity of the proposed approach concerning the number of layers, dimensions, and training rounds.
            }
	\label{fig:micro-F11}
    \end{flushleft}
\end{figure*}

\begin{figure*}[!h]
    \centering
    \begin{flushleft}
	\begin{subfigure}[t]{0.32\linewidth}
		\includegraphics[width=\textwidth]{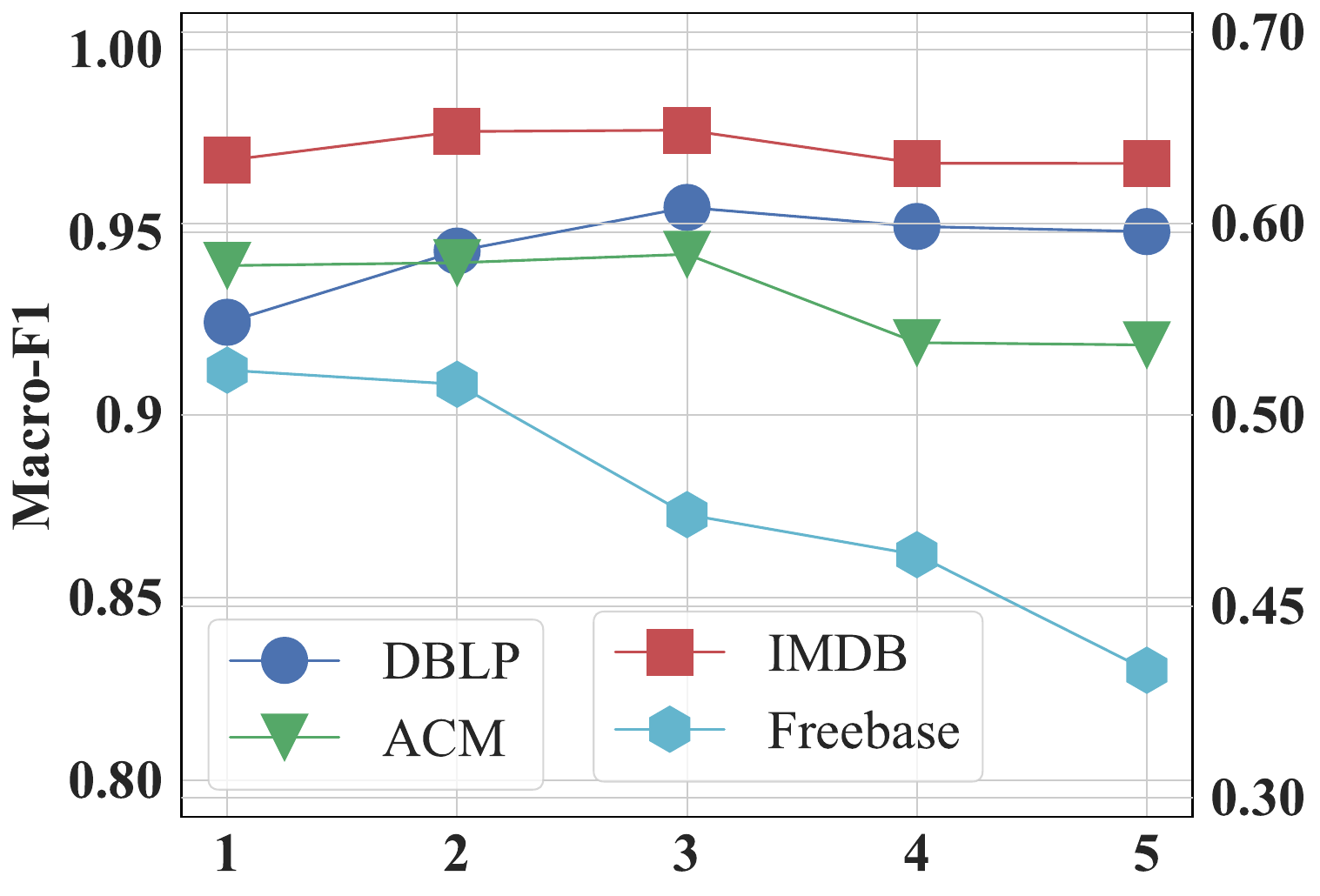}
		\caption{Macro-F1 score \emph{w.r.t.} \#layers}
		\label{fig:layers2}
	\end{subfigure}
	\begin{subfigure}[t]{0.32\linewidth}
		\includegraphics[width=\textwidth]{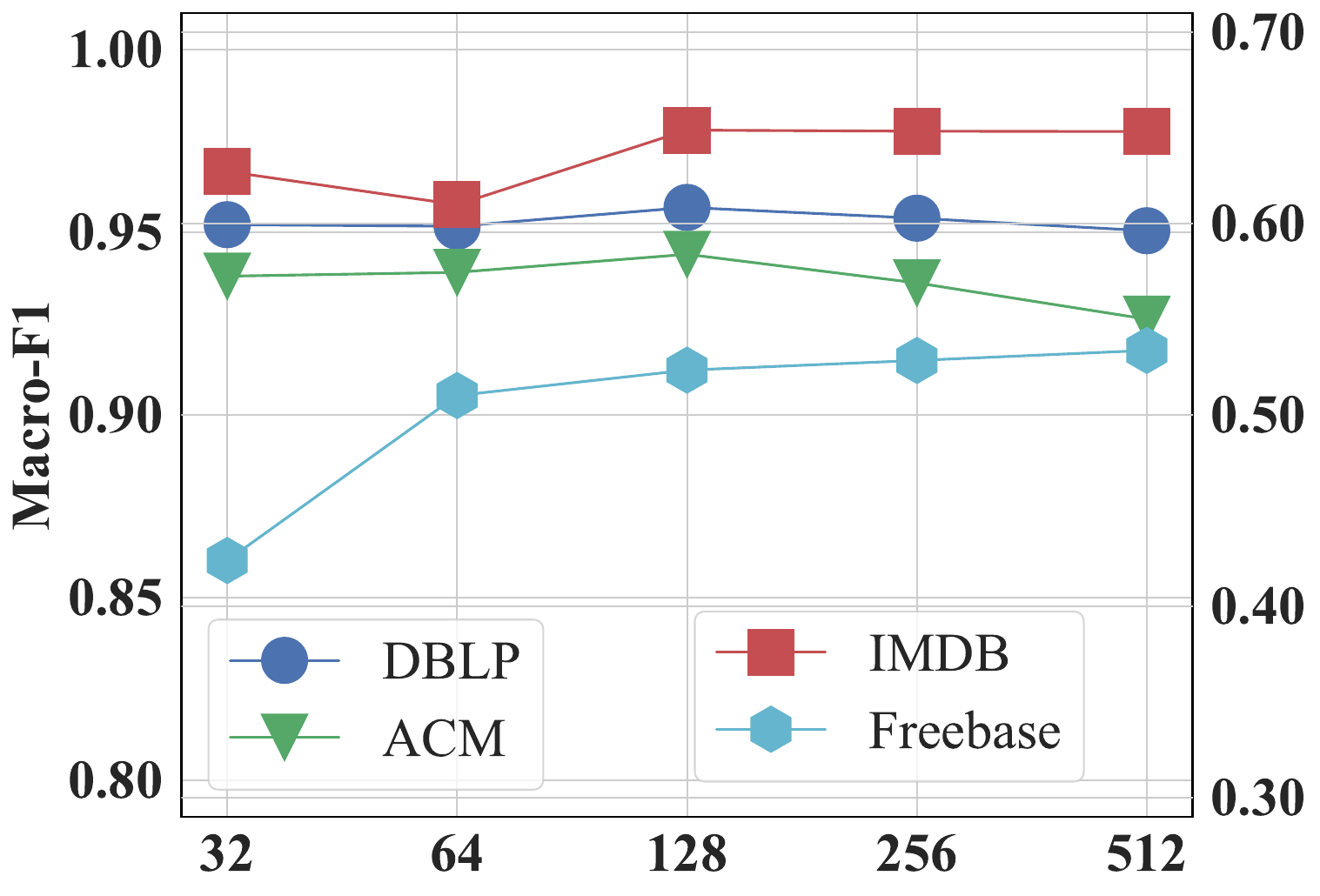}
		\caption{Macro-F1 score \emph{w.r.t.} \#dimensions}
		\label{fig:dim2}
	\end{subfigure}
	\begin{subfigure}[t]{0.32\linewidth}
		\includegraphics[width=\textwidth]{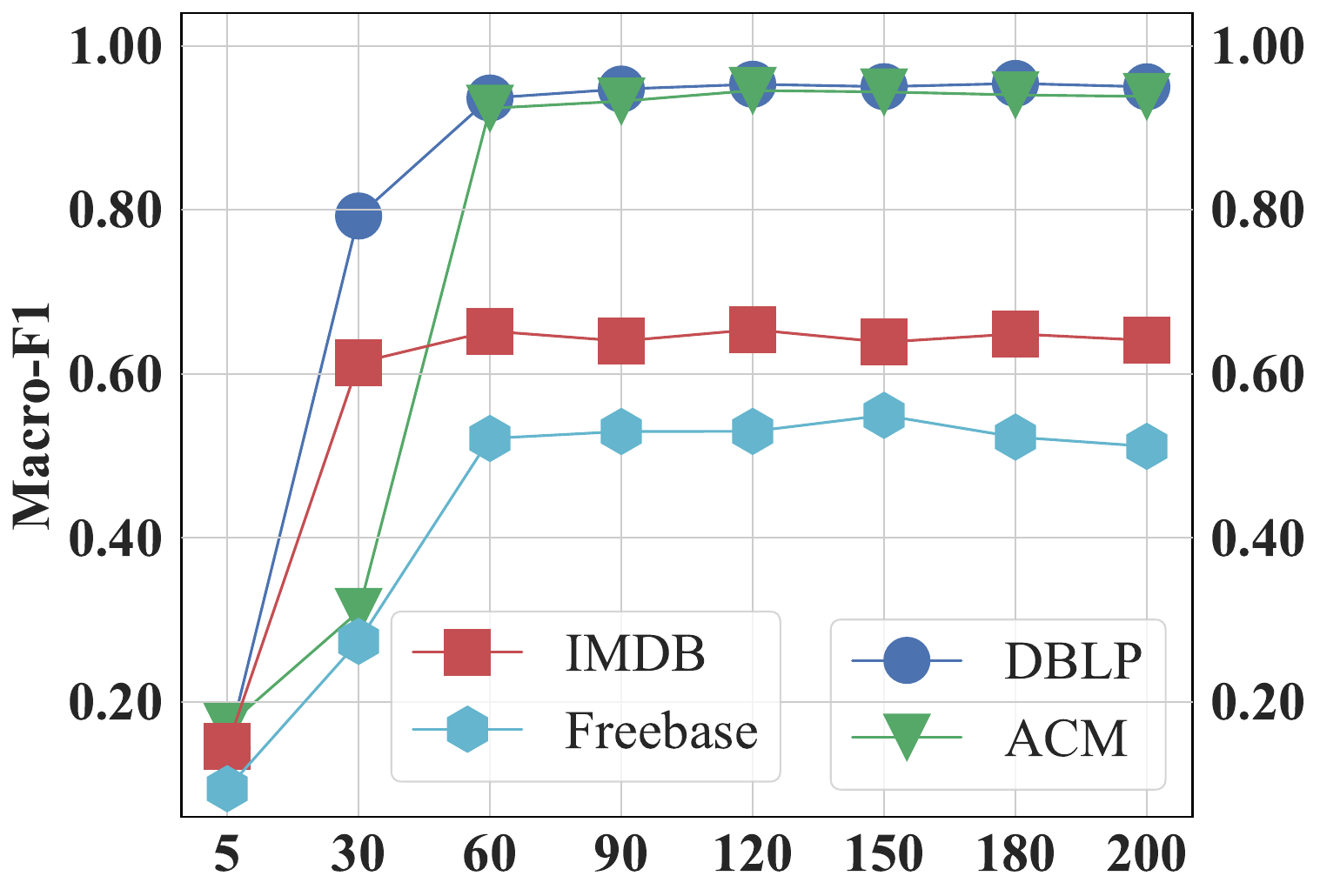}
		\caption{Macro-F1 score \emph{w.r.t.} \#rounds}
		\label{fig:rounds2}
	\end{subfigure}
	\caption{
        Macro-F1 outcomes depicting the parameter sensitivity of the proposed approach concerning the number of layers, dimensions, and training rounds.
        }
	\label{fig:micro-f2}
    \end{flushleft}
\end{figure*}

\subsection{Ablation Study (\textbf{RQ3})} \label{sec:ablation}
To substantiate the effectiveness of each element within our model, we proceed to conduct experiments involving different iterations of the \alg~framework.
\begin{itemize}[leftmargin=*]
    \item \emph{w/o} FGL: without mapping the edge feature representations to the node feature space and performing the dot product in the same feature space; 
    \item \emph{w/o} $L_{2}$: without considering $L_{2}$ normalization of the output by encoding the final stage of uniform mapping of nodes; 
    \item \emph{w/o} non-linear encoding (NLE): not considering a non-linear encoding mapping of nodes, as it only employs a fully connected layer for linear transformation; 
    \item \emph{w/o} edge information (EI): use random initialization to encode edges instead of using initialized encoding based on different edge types;
\end{itemize}
In Fig.~\ref{fig:ablation33}, we present the outcomes obtained through a comprehensive breakdown analysis conducted on four distinct datasets, focusing on the task of node classification. 
The performance metrics for IMDB and Freebase are aligned with the right-hand axis for reference.

The results demonstrate that both essential components contribute to the performance enhancement of our. 
In particular, \emph{w/o} FGL and \emph{w/o} EI have a substantial effect on the effectiveness of our model~\alg. 
The comparisons between \emph{w/o} FGL and \alg~highlight the effectiveness of edge-unified mapping encoding. 
We observe that \emph{w/o} FGL performs worse than \alg~in terms of macro-F1 and micro-F1, showing a decrease of 3.53\% in macro-F1 score on the DBLP dataset. 
This indicates that the designed unified encoding module effectively captures the crucial role of different relationship importance aspects in graph representation learning. 
\alg~outperforms \emph{w/o} FGL by 3.66\% and 1.16\% in terms of macro-F1 on DBLP and Freebase, respectively. 
It is possible that edge features and node features are fundamentally heterogeneous, and if message passing is performed directly, the edge features, serving as the channels for message passing, cannot effectively scale and learn useful features from neighboring nodes because they exist in different data spaces.
The comparisons between \emph{w/o} $L_{2}$ and \alg~highlight the significance of $L_{2}$ normalization on the output embedding.

Compared to \emph{w/o} NLE, \alg~improves by 0.53\% and 4.11\% in terms of micro-F1 on ACM and IMDB, respectively. 
Compared to \emph{w/o} EI, \alg~achieves an improvement of 3.86\% and 2.79\% in terms of macro-F1 on DBLP and ACM, respectively. 
For the initialization of the edge feature encoding, a random initialization was used instead of a type-based initialization for \emph{w/o} EI. 
This may lead to insufficient utilization of the type information of edges in the dataset during message passing, resulting in different types of edges receiving approximately distributed feature preferences. 
This can introduce noise when aggregating nodes, as unnecessary information is transmitted. 
For example, in a citation network, there is a citation relationship between papers, while there is a writing relationship between authors and papers. 
If paper obtains information about neighboring papers through encoding based on the writing action type, redundant feature information will inevitably be obtained.

\subsection{Time Analysis and Parameter Sensitivity (\textbf{RQ4})} \label{sec:time}
\begin{figure}[!t]
    \centering
    \includegraphics[width=0.45\textwidth]{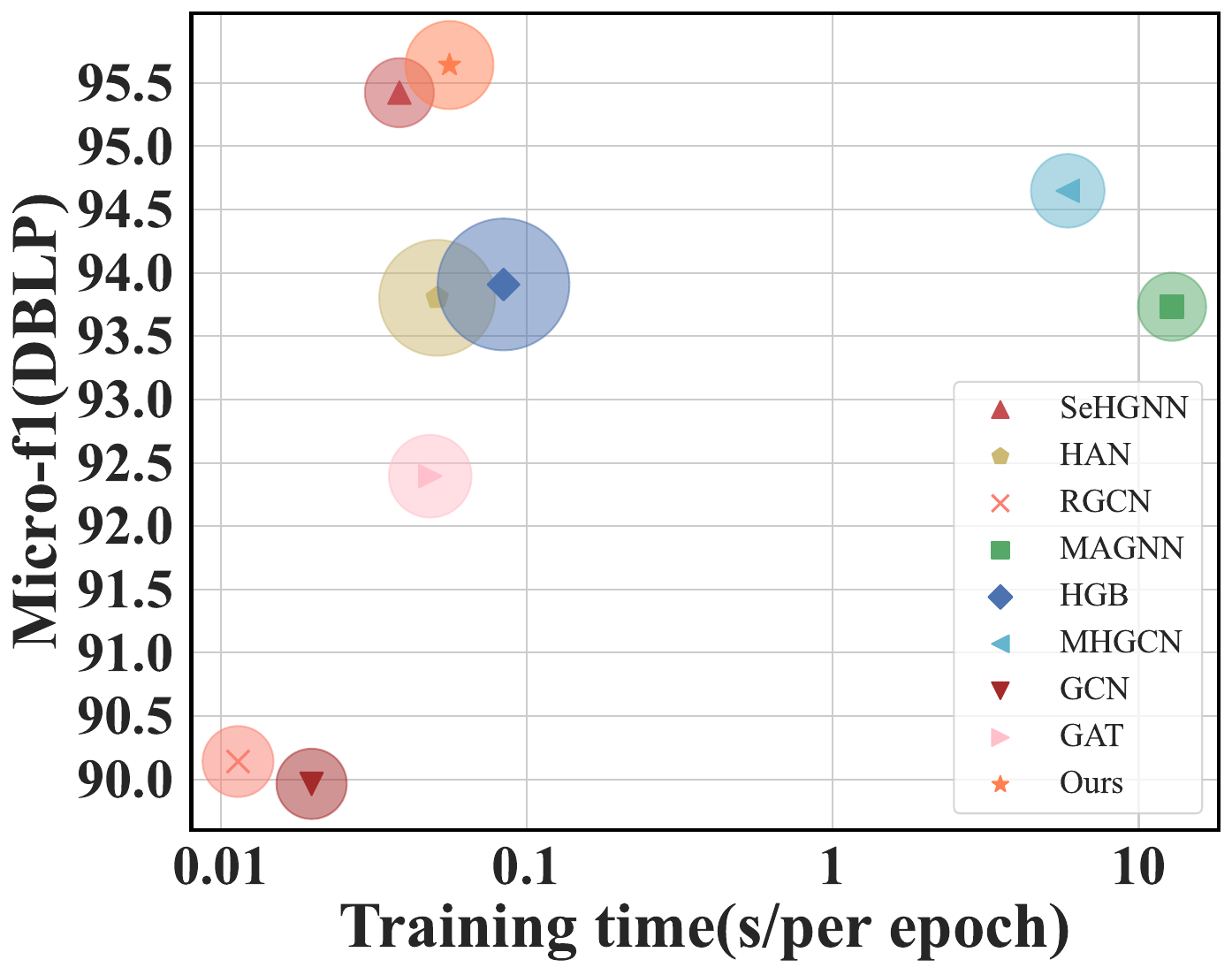}
    \caption{
    Contrast in time and memory consumption among HGNNs on the DBLP dataset. Circle sizes symbolize the (proportional) memory usage of the respective models.
    }
    \label{fig:time}
\end{figure}

{\bf Time Analysis.}
We evaluated the time and memory requirements of all accessible algorithms for node classification on the DBLP dataset. 
As observed in Fig. \ref{fig:time}, our proposed EdgeFGL model stands out in terms of Micro-f1 score, achieving an impressive score close to 95.5, significantly outperforming all other compared models. 
This demonstrates that our EdgeFGL can provide higher accuracy and more reliable results. 
The SeHGNN and HAN models exhibit a balanced performance in terms of micro-f1 score and training time, with high scores and low memory consumption. 
The RGCN, MAGNN, and HGB models achieve moderate micro-f1 scores, while the GCN and GAT models have relatively lower scores but require less training time and memory. 
Overall, our EdgeFGL effectively balances time and memory costs to achieve superior model performance. 


{\bf Parameter Sensitivity.}
We explore the \alg~sensitivity in relation to crucial parameters, including the number of layers $l$, embedding dimension $d$, and the quantity of training iterations.
Fig. \ref{fig:micro-F11} displays the macro-F1 score for node classification with various parameter settings across four datasets. 
Observe that the performance on FreeBase and IMDB corresponds to the rightmost ordinate.
As shown in Fig. \ref{fig:layers}, the performance of \alg~increases as $l$ increases, and then starts to decrease or level off when $l \geq $ 2. 
As the number of GCN layers increases, the representation of the nodes flattens out after multiple convolutions, resulting in a decrease in performance.
The results in Fig. \ref{fig:dim} show that the performance of \alg~gradually increases and then slightly decreases as the dimension $d$ increases, with DBLP and ACM reaching the best performance at embedding dimension $d$ = 128. 
%
This phenomenon arises due to the fact that when the dimension $d$ is small, the features of all nodes become compressed into a limited embedding space. Consequently, maintaining the feature proximity of all pairs of nodes becomes challenging.
Conversely, a higher dimensionality also tends to reduce the differentiation among all node embeddings. 
%
Figs. \ref{fig:rounds} and \ref{fig:rounds2} showcases the performance trend of our \alg~concerning the number of training rounds required for learning the model weights. 
It is evident that our \alg~can rapidly converge and consistently attain stable performance within 60 rounds across all evaluated datasets. 

\section{Conclusions}
\label{sec:Conclude}


The message-passing framework leverages the propagation characteristic of graph-structured data to enrich the information of each node, which makes it more suitable for node representation learning than traditional methods. 
In this paper, we have developed a novel message-passing based model, namely Edge-empowered Graph Feature Preference Learning (\alg), by integrating the edge representation learning into the messagepassing model, which can help refine the information in finer granularity for nodes. 
By doing this, nodes can receive more useful information, and thus promote the node representation learning process. 
Extensive experiments are conducted in both of node classification and node clustering on four datasets. 
And the results demonstrate the superior performance of node representations learned by \alg\, and the visualization results further reflect that the learned distribution can be better distributed in latent space. 
%


%
%
%

\bibliographystyle{IEEEtran}

\section*{Acknowledgement}
This work was supported by the Jinan University, 
National Natural Science Foundation of China (No.62272198),
Guangdong Key Laboratory of Data Security and Privacy Preserving (Grant No. 2023B1212060036),
Guangdong-Hong Kong Joint Laboratory for Data Security and Privacy Preserving (Grant No. 2023B1212120007),
Guangdong Basic and Applied Basic Research Foundation (No. 2024A1515010121), and
the Outstanding Innovative Talents Cultivation Funded Programs for Doctoral Students of Jinan University (No.2023CXB022).


\bibliography{ref.bib}

\begin{thebibliography}{10}
\providecommand{\url}[1]{#1}
\csname url@samestyle\endcsname
\providecommand{\newblock}{\relax}
\providecommand{\bibinfo}[2]{#2}
\providecommand{\BIBentrySTDinterwordspacing}{\spaceskip=0pt\relax}
\providecommand{\BIBentryALTinterwordstretchfactor}{4}
\providecommand{\BIBentryALTinterwordspacing}{\spaceskip=\fontdimen2\font plus
\BIBentryALTinterwordstretchfactor\fontdimen3\font minus \fontdimen4\font\relax}
\providecommand{\BIBforeignlanguage}[2]{{%
\expandafter\ifx\csname l@#1\endcsname\relax
\typeout{** WARNING: IEEEtran.bst: No hyphenation pattern has been}%
\typeout{** loaded for the language `#1'. Using the pattern for}%
\typeout{** the default language instead.}%
\else
\language=\csname l@#1\endcsname
\fi
#2}}
\providecommand{\BIBdecl}{\relax}
\BIBdecl

\bibitem{caos2015learninggraphrepresentations}
L.~CAOS and Q.~G. XU, ``Learning graph representations with global structural information,'' \emph{Proceedings of CIKM}, 2015.

\bibitem{du2021tabularnet}
L.~Du, F.~Gao, X.~Chen, R.~Jia, J.~Wang, J.~Zhang, S.~Han, and D.~Zhang, ``Tabularnet: A neural network architecture for understanding semantic structures of tabular data,'' in \emph{Proceedings of KDD}, 2021, pp. 322--331.

\bibitem{yu2020structured}
Z.~Yu, Z.~Zhang, H.~Chen, and J.~Shao, ``Structured subspace embedding on attributed networks,'' \emph{Information Sciences}, vol. 512, pp. 726--740, 2020.

\bibitem{wang2016structural}
D.~Wang, P.~Cui, and W.~Zhu, ``Structural deep network embedding,'' in \emph{Proceedings of KDD}, 2016, pp. 1225--1234.

\bibitem{cao2016deep}
S.~Cao, W.~Lu, and Q.~Xu, ``Deep neural networks for learning graph representations,'' in \emph{Proceedings of AAAI}, vol.~30, no.~1, 2016.

\bibitem{he2019content}
Z.~He, J.~Liu, Y.~Zeng, L.~Wei, and Y.~Huang, ``Content to node: Self-translation network embedding,'' \emph{IEEE Transactions on Knowledge and Data Engineering}, vol.~33, no.~2, pp. 431--443, 2019.

\bibitem{velivckovic2017graph}
P.~Veli{\v{c}}kovi{\'c}, G.~Cucurull, A.~Casanova, A.~Romero, P.~Lio, and Y.~Bengio, ``Graph attention networks,'' \emph{arXiv preprint arXiv:1710.10903}, 2017.

\bibitem{lipton2015critical}
Z.~C. Lipton, J.~Berkowitz, and C.~Elkan, ``A critical review of recurrent neural networks for sequence learning,'' \emph{arXiv preprint arXiv:1506.00019}, 2015.

\bibitem{shi2022heterogeneous}
C.~Shi, ``Heterogeneous graph neural networks,'' \emph{Graph Neural Networks: Foundations, Frontiers, and Applications}, pp. 351--369, 2022.

\bibitem{wang2019heterogeneous}
X.~Wang, H.~Ji, C.~Shi, B.~Wang, Y.~Ye, P.~Cui, and P.~S. Yu, ``Heterogeneous graph attention network,'' in \emph{The world wide web conference}, 2019, pp. 2022--2032.

\bibitem{xu2022explicit}
L.~Xu, Z.-Y. He, K.~Wang, C.-D. Wang, and S.-Q. Huang, ``Explicit message-passing heterogeneous graph neural network,'' \emph{IEEE Transactions on Knowledge and Data Engineering}, 2022.

\bibitem{ding2021diffmg}
Y.~Ding, Q.~Yao, H.~Zhao, and T.~Zhang, ``Diffmg: Differentiable meta graph search for heterogeneous graph neural networks,'' in \emph{Proceedings of KDD}, 2021, pp. 279--288.

\bibitem{kipf2016semi}
T.~N. Kipf and M.~Welling, ``Semi-supervised classification with graph convolutional networks,'' \emph{arXiv preprint arXiv:1609.02907}, 2016.

\bibitem{yu2022multiplex}
P.~Yu, C.~Fu, Y.~Yu, C.~Huang, Z.~Zhao, and J.~Dong, ``Multiplex heterogeneous graph convolutional network,'' in \emph{Proceedings of KDD}, 2022, pp. 2377--2387.

\bibitem{goldberg2014word2vec}
Y.~Goldberg and O.~Levy, ``word2vec explained: deriving mikolov et al.'s negative-sampling word-embedding method,'' \emph{arXiv preprint arXiv:1402.3722}, 2014.

\bibitem{roweis2000nonlinear}
S.~T. Roweis and L.~K. Saul, ``Nonlinear dimensionality reduction by locally linear embedding,'' \emph{science}, vol. 290, no. 5500, pp. 2323--2326, 2000.

\bibitem{tenenbaum2000global}
J.~B. Tenenbaum, V.~d. Silva, and J.~C. Langford, ``A global geometric framework for nonlinear dimensionality reduction,'' \emph{science}, vol. 290, no. 5500, pp. 2319--2323, 2000.

\bibitem{belkin2001laplacian}
M.~Belkin and P.~Niyogi, ``Laplacian eigenmaps and spectral techniques for embedding and clustering,'' \emph{Proceedings of NeurIPS}, vol.~14, 2001.

\bibitem{mikolov2013distributed}
T.~Mikolov, I.~Sutskever, K.~Chen, G.~S. Corrado, and J.~Dean, ``Distributed representations of words and phrases and their compositionality,'' \emph{Proceedings of NeurIPS}, vol.~26, 2013.

\bibitem{perozzi2014deepwalk}
B.~Perozzi, R.~Al-Rfou, and S.~Skiena, ``Deepwalk: Online learning of social representations,'' in \emph{Proceedings of KDD}, 2014, pp. 701--710.

\bibitem{grover2016node2vec}
A.~Grover and J.~Leskovec, ``node2vec: Scalable feature learning for networks,'' in \emph{Proceedings of KDD}, 2016, pp. 855--864.

\bibitem{wang2017community}
X.~Wang, P.~Cui, J.~Wang, J.~Pei, W.~Zhu, and S.~Yang, ``Community preserving network embedding,'' in \emph{Proceedings of AAAI}, vol.~31, no.~1, 2017.

\bibitem{li2021attributed}
J.-H. Li, L.~Huang, C.-D. Wang, D.~Huang, J.-H. Lai, and P.~Chen, ``Attributed network embedding with micro-meso structure,'' \emph{ACM Transactions on Knowledge Discovery from Data}, vol.~15, no.~4, pp. 1--26, 2021.

\bibitem{bullinaria2007extracting}
J.~A. Bullinaria and J.~P. Levy, ``Extracting semantic representations from word co-occurrence statistics: A computational study,'' \emph{Behavior research methods}, vol.~39, pp. 510--526, 2007.

\bibitem{tu2017cane}
C.~Tu, H.~Liu, Z.~Liu, and M.~Sun, ``Cane: Context-aware network embedding for relation modeling,'' in \emph{Proceedings of ACL}, 2017, pp. 1722--1731.

\bibitem{liu2018content}
J.~Liu, Z.~He, L.~Wei, and Y.~Huang, ``Content to node: Self-translation network embedding,'' in \emph{Proceedings of KDD}, 2018, pp. 1794--1802.

\bibitem{gao2018deep}
H.~Gao and H.~Huang, ``Deep attributed network embedding,'' in \emph{Proceedings of IJCAI}, 2018.

\bibitem{gao2018self}
------, ``Self-paced network embedding,'' in \emph{Proceedings of KDD}, 2018, pp. 1406--1415.

\bibitem{shen2018deep}
X.~Shen and F.-L. Chung, ``Deep network embedding for graph representation learning in signed networks,'' \emph{IEEE transactions on cybernetics}, vol.~50, no.~4, pp. 1556--1568, 2018.

\bibitem{wang2020node}
C.-D. Wang, W.~Shi, L.~Huang, K.-Y. Lin, D.~Huang, and S.~Y. Philip, ``Node pair information preserving network embedding based on adversarial networks,'' \emph{IEEE Transactions on Cybernetics}, vol.~52, no.~7, pp. 5908--5922, 2020.

\bibitem{hu2019hierarchical}
F.~Hu, Y.~Zhu, S.~Wu, L.~Wang, and T.~Tan, ``Hierarchical graph convolutional networks for semi-supervised node classification,'' \emph{arXiv preprint arXiv:1902.06667}, 2019.

\bibitem{xu2019relation}
F.~Xu, J.~Lian, Z.~Han, Y.~Li, Y.~Xu, and X.~Xie, ``Relation-aware graph convolutional networks for agent-initiated social e-commerce recommendation,'' in \emph{Proceedings of CIKM}, 2019, pp. 529--538.

\bibitem{jiang2019dynamic}
J.~Jiang, Y.~Wei, Y.~Feng, J.~Cao, and Y.~Gao, ``Dynamic hypergraph neural networks.'' in \emph{Proceedings of IJCAI}, 2019, pp. 2635--2641.

\bibitem{hamilton2017inductive}
W.~Hamilton, Z.~Ying, and J.~Leskovec, ``Inductive representation learning on large graphs,'' \emph{Proceedings of NeurIPS}, vol.~30, 2017.

\bibitem{pan2019learning}
S.~Pan, R.~Hu, S.-f. Fung, G.~Long, J.~Jiang, and C.~Zhang, ``Learning graph embedding with adversarial training methods,'' \emph{IEEE transactions on cybernetics}, vol.~50, no.~6, pp. 2475--2487, 2019.

\bibitem{jiang2019censnet}
X.~Jiang, P.~Ji, and S.~Li, ``Censnet: Convolution with edge-node switching in graph neural networks.'' in \emph{Proceedings of IJCAI}, 2019, pp. 2656--2662.

\bibitem{ye2019vectorized}
R.~Ye, X.~Li, Y.~Fang, H.~Zang, and M.~Wang, ``A vectorized relational graph convolutional network for multi-relational network alignment.'' in \emph{Proceedings of IJCAI}, 2019, pp. 4135--4141.

\bibitem{he2020realformer}
R.~He, A.~Ravula, B.~Kanagal, and J.~Ainslie, ``Realformer: Transformer likes residual attention,'' \emph{arXiv preprint arXiv:2012.11747}, 2020.

\bibitem{bollacker2008freebase}
K.~Bollacker, C.~Evans, P.~Paritosh, T.~Sturge, and J.~Taylor, ``Freebase: a collaboratively created graph database for structuring human knowledge,'' in \emph{Proceedings of SIGMOD}, 2008, pp. 1247--1250.

\bibitem{yang2020heterogeneous}
C.~Yang, Y.~Xiao, Y.~Zhang, Y.~Sun, and J.~Han, ``Heterogeneous network representation learning: A unified framework with survey and benchmark,'' \emph{IEEE Transactions on Knowledge and Data Engineering}, vol.~34, no.~10, pp. 4854--4873, 2020.

\bibitem{yun2019graph}
S.~Yun, M.~Jeong, R.~Kim, J.~Kang, and H.~J. Kim, ``Graph transformer networks,'' \emph{Proceedings of NeurIPS}, vol.~32, 2019.

\bibitem{hu2020heterogeneous}
Z.~Hu, Y.~Dong, K.~Wang, and Y.~Sun, ``Heterogeneous graph transformer,'' in \emph{Proceedings of WWW}, 2020, pp. 2704--2710.

\bibitem{schlichtkrull2018modeling}
M.~Schlichtkrull, T.~N. Kipf, P.~Bloem, R.~Van Den~Berg, I.~Titov, and M.~Welling, ``Modeling relational data with graph convolutional networks,'' in \emph{Proceedings of ESWC}.\hskip 1em plus 0.5em minus 0.4em\relax Springer, 2018, pp. 593--607.

\bibitem{fu2020magnn}
X.~Fu, J.~Zhang, Z.~Meng, and I.~King, ``Magnn: Metapath aggregated graph neural network for heterogeneous graph embedding,'' in \emph{Proceedings of WWW}, 2020, pp. 2331--2341.

\bibitem{lv2021we}
Q.~Lv, M.~Ding, Q.~Liu, Y.~Chen, W.~Feng, S.~He, C.~Zhou, J.~Jiang, Y.~Dong, and J.~Tang, ``Are we really making much progress? revisiting, benchmarking and refining heterogeneous graph neural networks,'' in \emph{Proceedings of KDD}, 2021, pp. 1150--1160.

\bibitem{yang2023simple}
X.~Yang, M.~Yan, S.~Pan, X.~Ye, and D.~Fan, ``Simple and efficient heterogeneous graph neural network,'' in \emph{Proceedings of AAAI}, vol.~37, no.~9, 2023, pp. 10\,816--10\,824.

\bibitem{li2022multi}
Z.~Li, Q.~Ren, L.~Chen, X.~Sui, and J.~Li, ``Multi-hierarchical spatial-temporal graph convolutional networks for traffic flow forecasting,'' in \emph{Proceedings of ICPR}.\hskip 1em plus 0.5em minus 0.4em\relax IEEE, 2022, pp. 4913--4919.

\bibitem{liao2018attributed}
L.~Liao, X.~He, H.~Zhang, and T.-S. Chua, ``Attributed social network embedding,'' \emph{IEEE Transactions on Knowledge and Data Engineering}, vol.~30, no.~12, pp. 2257--2270, 2018.

\end{thebibliography}

\begin{IEEEbiography}    
  [{\includegraphics[width=1in,height=1.25in,
  clip,keepaspectratio]
  {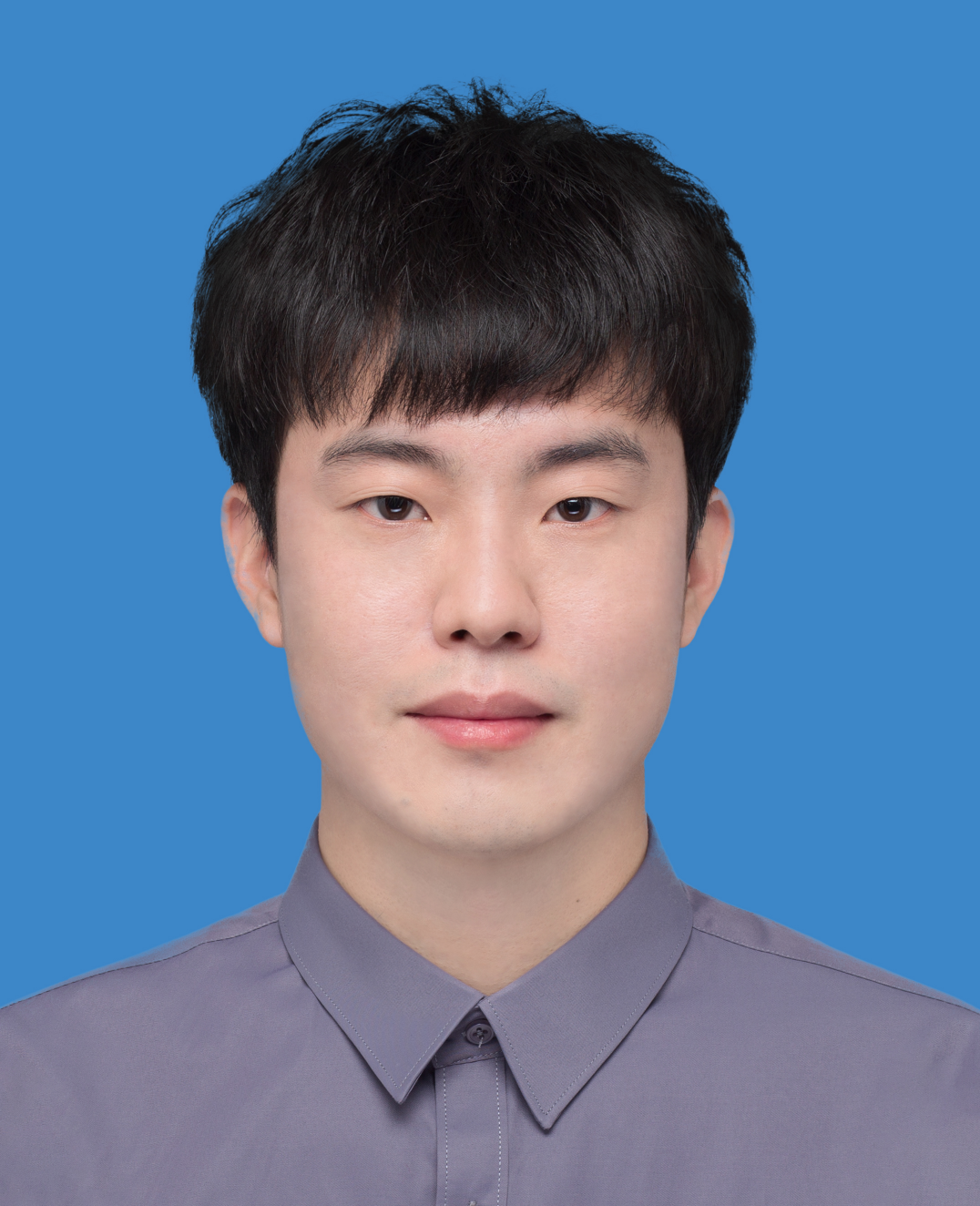}}]{Shengda Zhuo} received the M.S. degree in computer technology from Guangzhou 
  University, Guangzhou, China in 2023. He is currently pursuing the Ph.D. degree in Cyberspace Security at Jinan University, Guangzhou, China. 
  His research interests include graph neural networks, recommendation systems and data mining. 
\end{IEEEbiography}

\begin{IEEEbiography}    
  [{\includegraphics[width=1in,height=1.25in,
  clip,keepaspectratio]
  {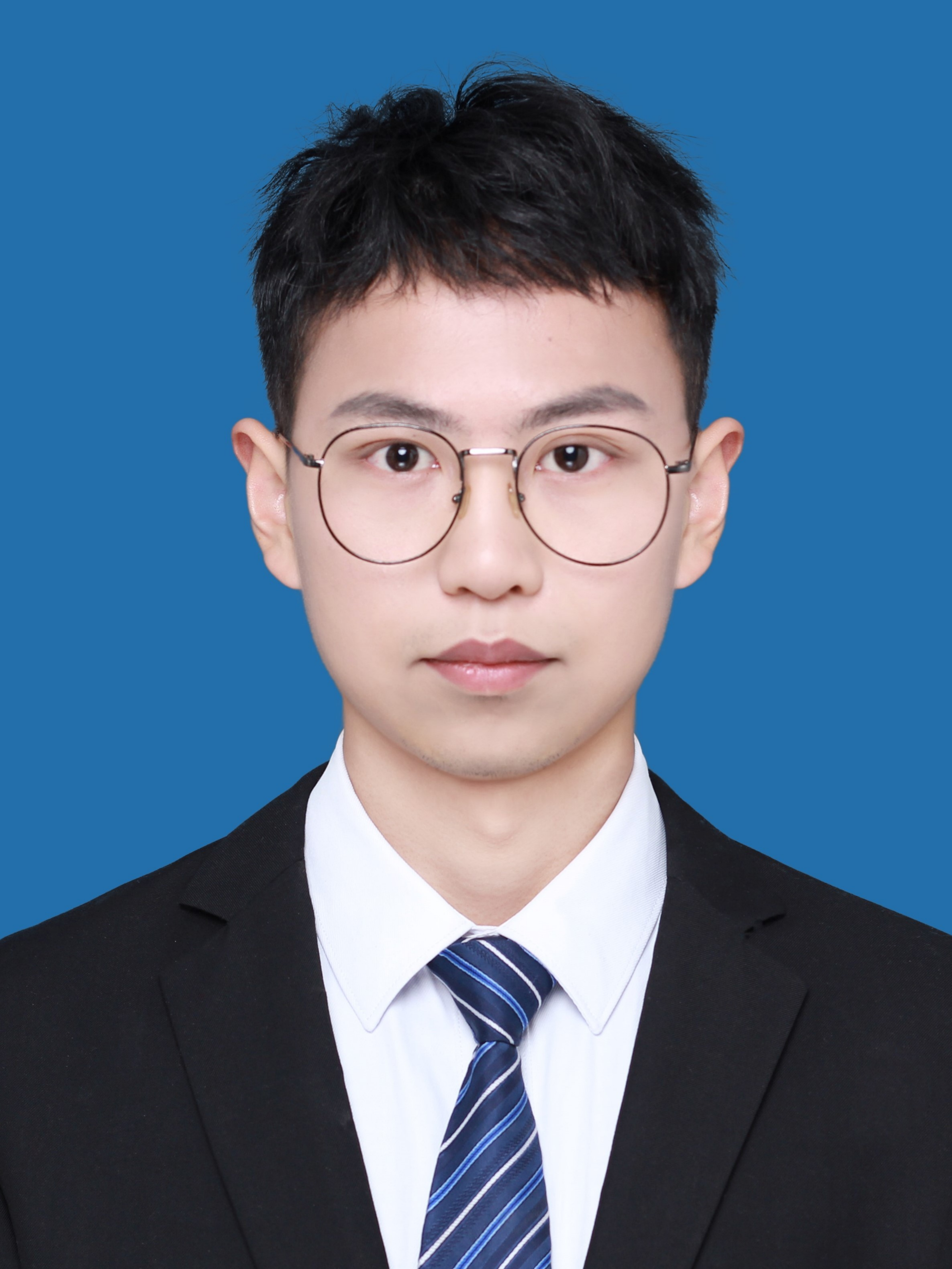}}]{Ji-Wang Fang} received his bachelor's degree in 2021 from Anhui Normal University.  He is currently pursuing the M.S degree in Institute of Technology at Jinan University.  Guangzhou, China. His main research focuses on graph neural networks.
\end{IEEEbiography}

\begin{IEEEbiography}    
  [{\includegraphics[width=1in,height=1.25in,
  clip,keepaspectratio]
  {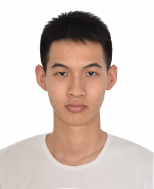}}]{Hong-Guang Lin} received his bachelor's degree in 2021 from Shantou University. He is currently pursuing the M.S degree in cyberspace security at Jinan University. Guangzhou, China. His main research focuses on
  artificial intelligence.
\end{IEEEbiography}

\begin{IEEEbiography}[{\includegraphics[width=1in,height=1.25in,clip,keepaspectratio]{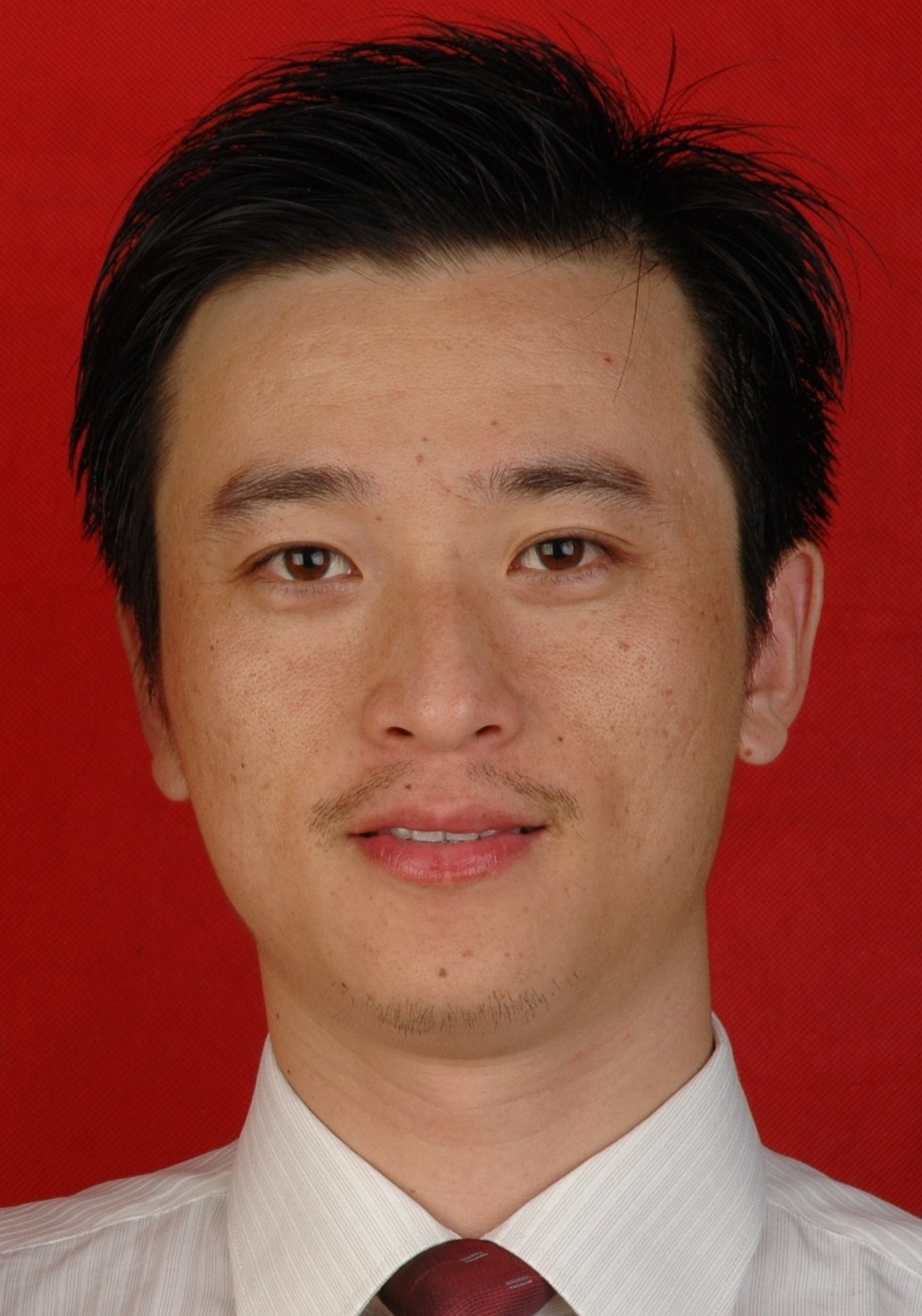}}]{Yin Tang} obtained his Ph.D. from the Laboratory of Machine Learning and Data Mining at the School of Computer Science and Engineering, South China University of Technology in 2004, and currently is a professor at Jinan University, serving as the deputy director of the National Experimental Center for Economics and Management (Jinan University). His research interests cover the fields of the Big Data and Intelligent Business, Enterprise Brain with AI, and Computational Social Science. He has published over 50 papers in international journals and conferences, and has been granted over 10 patents. 
\end{IEEEbiography}


\begin{IEEEbiography}
  [{\includegraphics[width=1in,height=1.25in,clip,keepaspectratio]{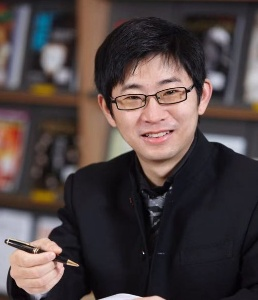}}]{Min Chen}
  Currently serving as a professor and doctoral supervisor at the School of Computer Science and Engineering, South China University of Technology, a distinguished faculty member as an IEEE Fellow and an IET Fellow. With over 39,500 citations on Google Scholar and an H-index of 94, he has been recognized as a highly cited researcher by Clarivate Analytics in 2018, 2019, 2020, 2021, and 2022.
  
  Completing his doctoral studies at the School of Electronic and Communication Engineering, South China University of Technology at the age of 23, he went on to pursue his postdoctoral research at Seoul National University, South Korea, and the University of British Columbia, Canada. In 2009, he joined Seoul National University as a faculty member and returned to China in 2012 to establish the Embedded and Pervasive Computing Laboratory at Huazhong University of Science and Technology. Currently, he holds the position of professor and doctoral supervisor at the School of Computer Science and Engineering, South China University of Technology.
  
  He has published over 200 papers in internationally reputed journals and conferences, including IEEE JSAC, IEEE TNNLS, IEEE TPDS, IEEE TWC, IEEE TSC, INFOCOM, Science, Nature Communications, and more. Furthermore, he has been granted over 20 national invention patents and authored 12 books and educational materials. Notably, his English textbook "Applications of Big Data Analytics" has been adopted by prestigious institutions such as Harvard University and Stanford University.
  
  As an esteemed researcher in his field, he has been invited to deliver presentations at 16 international academic conferences. Several of his papers have received the Best Paper Awards, including the IEEE Communications Society Fred W. Ellersick Prize in 2017, the IEEE Vehicular Technology Society Jack Neubauer Memorial Award in 2019, and the Best Paper Award in the IEEE ComSoc Asia Pacific Region in 2022.
  \end{IEEEbiography}

  \begin{IEEEbiography}    
    [{\includegraphics[width=1in,height=1.25in,
    clip,keepaspectratio]
    {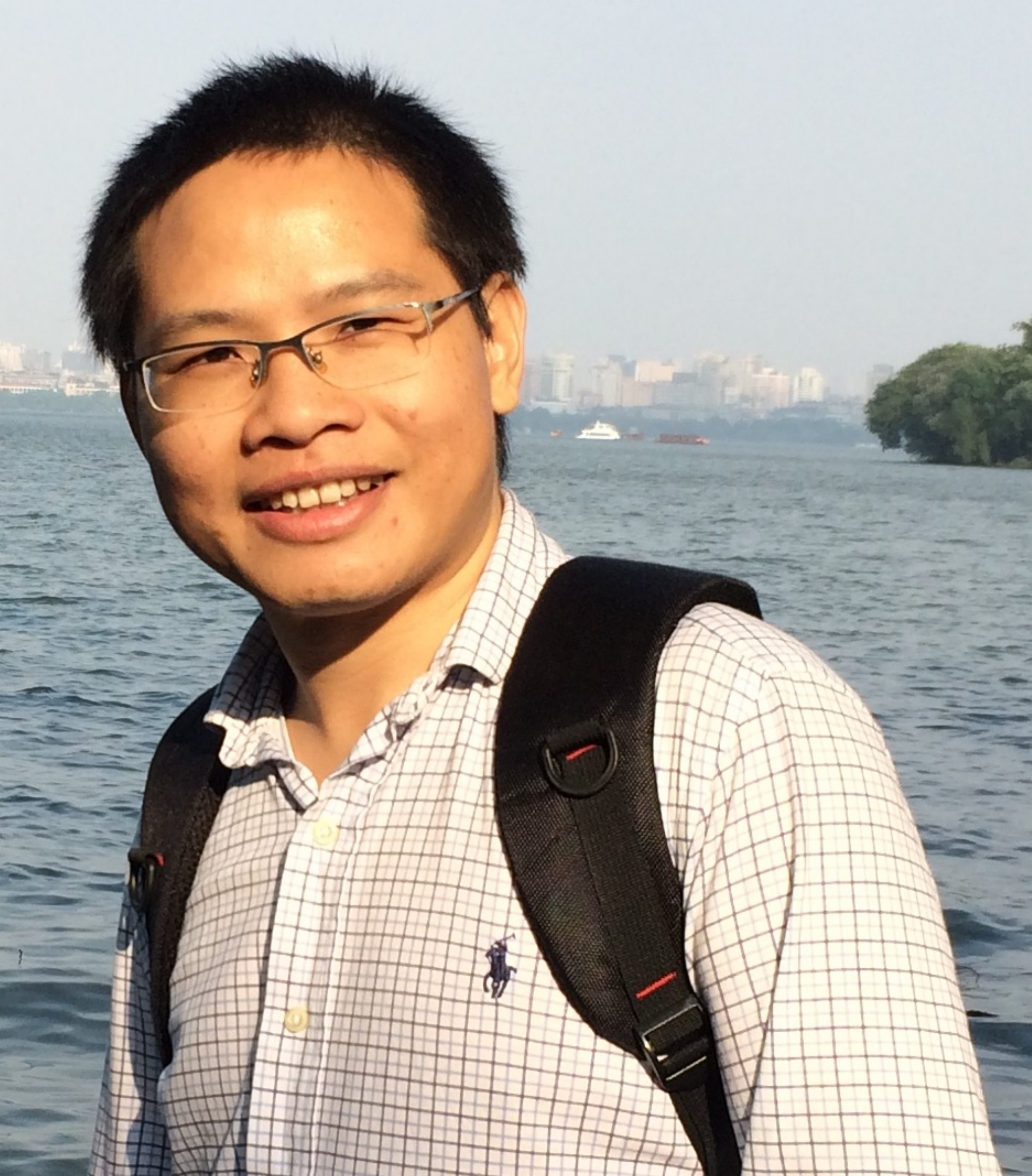}}]{Chang-Dong Wang} (Senior Member, IEEE) received the Ph.D. degree in computer science in 2013 from Sun Yat-sen University, Guangzhou, China. He is a visiting student at University of Illinois at Chicago from Jan. 2012 to Nov. 2012. He joined Sun Yat-sen University in 2013 as an assistant professor with School of Mobile Information Engineering and now he is currently an associate professor with School of Computer. His current research interests include machine learning and data mining. He has published over 120 scientific papers in international journals and conferences such as IEEE TPAMI, IEEE TCYB, IEEE TKDE, IEEE TNNLS, ACM TKDD, IEEE TSMC-Systems, IEEE TSMC-C, KDD, AAAI, IJCAI, CVPR, ICDM, CIKM and SDM. 
    
    His ICDM 2010 paper won the Honorable Mention for Best Research Paper Awards. He won 2012 Microsoft Research Fellowship Nomination Award. He was awarded 2015 Chinese Association for Artificial Intelligence (CAAI) Outstanding Dissertation. He is PC Co-chair of the 16th International Conference on Advanced Data Mining and Applications (ADMA 2020). He is an Associate Editor in Journal of Artificial Intelligence Research (JAIR).
  \end{IEEEbiography}
  
\begin{IEEEbiography}    
  [{\includegraphics[width=1in,height=1.25in,
  clip,keepaspectratio]
  {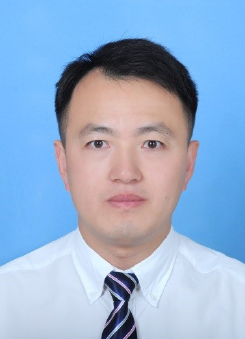}}]{Shu-Qiang Huang} received the Ph.D. degree in computer science in 2010 from South China University of Technology, Guangzhou, China. He is currently a full professor of college of cyber security of Jinan University. He is a distinguished member of China Computer Federation. His main research interests include edge computing, industrial IoT and artificial intelligence. He has published more than 50 academic papers in international journals such as IEEE TCYB, IEEE TPDS, IEEE TKDE, IEEE TII, TCS, and ACM TIST.
\end{IEEEbiography}

\end{document}